\documentclass[10pt,twocolumn,letterpaper]{article}

\usepackage{cvpr}
\usepackage{times}
\usepackage{epsfig}
\usepackage{graphicx}
\usepackage{amsmath}
\usepackage{amssymb}
\usepackage{algorithm}
\usepackage{algorithmic}

\usepackage{subfig} 
\usepackage{pst-plot}
\usepackage{pstricks-add}    

\usepackage[pagebackref=true,breaklinks=true,letterpaper=true,colorlinks,bookmarks=false]{hyperref}

\def\x{{\mathbf x}}
\def\z{{\mathbf z}}
\def\1{{\mathbf 1}}

\def\q{{\mathbf q}}

\def\X{{\mathbf X}}

\def\Z{{\mathbf Z}}

\def\Gammab{{\boldsymbol\Gamma}}

\def\betab{{\boldsymbol\beta}}

\def\alphab{{\boldsymbol\alpha}}

\def\Gammab{{\boldsymbol\Gamma}}

\def\w{{\mathbf w}}

\def\B{{\mathbf B}}
\def\Q{{\mathbf Q}}

\def\R{{\mathbf R}}

\def\Real{{\mathbb R}}

\def\A{{\mathbf A}}
\def\AA{{\mathcal A}}

\def\argmin{\operatornamewithlimits{arg\,min}}

\def\diag{\operatorname{diag}}

\def\st{~~\text{s.t.}~~}
\def\defin{\triangleq}

\long\def\symbolfootnote[#1]#2{\begingroup\def\thefootnote{\fnsymbol{footnote}}\footnote[#1]{#2}\endgroup}

\def\begincondeq{\ifthenelse{\isundefined{\supplemental}}{
\begin{displaymath}
}{
\begin{equation}
}}
\def\endcondeq{\ifthenelse{\isundefined{\supplemental}}{
\end{displaymath}
}{
\end{equation}
}}
\def\condvspace{\ifthenelse{\isundefined{\supplemental}}{\vspace*{-0.1cm}}{}}
\def\condvspacesmall{\ifthenelse{\isundefined{\supplemental}}{\vspace*{-0.05cm}}{}}
\newcommand{\myvspace}[1]{\ifthenelse{\isundefined{\supplemental}}{\vspace*{-#1cm}}{}}

\cvprfinalcopy 

\def\cvprPaperID{1331} 

\ifcvprfinal\fi
\begin{document}

\title{Fast and Robust Archetypal Analysis for Representation Learning}

\author{Yuansi Chen$^{1,2}$, \;\; Julien Mairal$^{2}$, \;\; Zaid Harchaoui$^{2}$\\
   $^1$EECS Department, University of California, Berkeley, \;\;$^2$Inria$^*$ \\
$^1${\tt\small yuansi.chen@berkeley.edu}, \;\;$^2${\tt\small firstname.lastname@inria.fr}
}
\maketitle
\renewcommand{\thefootnote}{\fnsymbol{footnote}}
\footnotetext[1]{LEAR team, Inria Grenoble Rh\^one-Alpes, Laboratoire Jean Kuntzmann, CNRS, Univ. Grenoble Alpes.}

\begin{abstract}
   We revisit a pioneer unsupervised learning technique called archetypal
analysis~\cite{Cut94}, which is related to successful data analysis methods
such as sparse coding~\cite{Ols96} and non-negative matrix
factorization~\cite{Paa94}. Since it was proposed, archetypal analysis did not
gain a lot of popularity even though it produces more interpretable models than
other alternatives. Because no efficient implementation has ever been made
publicly available, its application to important scientific problems may have been
severely limited.

Our goal is to bring back into favour archetypal analysis. We propose a fast
optimization scheme using an active-set strategy, and provide an efficient
open-source implementation interfaced with Matlab, R, and Python.  Then, we
demonstrate the usefulness of archetypal analysis for computer vision tasks,
such as codebook learning, signal classification, and large image
collection visualization.

\end{abstract}

\vspace*{-0.15cm}
\section{Introduction}

\vspace*{-0.05cm}
Unsupervised learning techniques have been widely used to automatically
discover the underlying structure of data. This may
serve several purposes, depending on the task considered. In experimental
sciences, one may be looking for data representations that automatically exhibit
interpretable patterns, for example groups of neurons with similar activation
in a population, clusters of genes manifesting similar
expression~\cite{Eis98}, or topics learned from text collections~\cite{Ble03}.

In image processing and computer vision, unsupervised learning is often used as
a data modeling step for a subsequent prediction task. For example, natural
image patches have been modeled with sparse coding~\cite{Ols96} or mixture of
Gaussians~\cite{Yu12}, yielding powerful representations for image restoration.
Similarly, local image descriptors have been encoded with unsupervised learning
methods~\cite{Csu04,Laz06,Yan09}, producing successful codebooks for visual
recognition pipelines.  Interpretation is probably not crucial for these
prediction tasks. However, it can be important for other purposes, \eg, for
data visualization.

Our main objective is to rehabilitate a pioneer unsupervised learning
technique called archetypal analysis~\cite{Cut94}, which is easy to interpret
while providing good results in prediction tasks. It was proposed
as an alternative to principal component analysis (PCA) for discovering latent
factors from high-dimensional data. Unlike
principal components, each factor learned by archetypal
analysis, called \emph{archetype}, is forced to be a convex combination of a
few data points. Such associations between archetypes and data points are 
useful for interpretation. For example, clustering techniques provide
such associations between data and centroids.
It is indeed common in genomics to cluster gene expression data
from several individuals, and to interpret each centroid by looking for some common
physiological traits among individuals of the same cluster~\cite{Eis98}.

Interestingly, archetypal analysis is related to popular approaches such as
sparse coding~\cite{Ols96} and non-negative matrix factorization
(NMF)~\cite{Paa94}, even though all these formulations were independently
invented around the same time. Archetypal analysis indeed produces sparse
representations of the data points, by approximating them with convex
combinations of archetypes; it also provides a non-negative factorization when
the data matrix is non-negative.

A natural question is why archetypal analysis did not gain a lot of
success, unlike NMF or sparse coding. We believe that the lack of efficient
available software has limited the deployment of archetypal analysis to promising
applications; our goal is to address this issue. First, we develop an
efficient optimization technique based on an active-set
strategy~\cite{Noc06}. Then, we demonstrate that our approach is scalable
and orders of magnitude faster than existing publicly available implementations.
Finally, we show that archetypal analysis can be useful for computer
vision, and we believe that it could have many applications in
other fields, such as neurosciences, bioinformatics, or natural
language processing.  We first show that it performs as well as sparse coding for
learning codebooks of features in visual recognition tasks~\cite{Yan09} and for
signal classification~\cite{Mai12,Ram10,Wri09}.
Second, we show that archetypal analysis provides a simple and effective way 
for visualizing large databases of images.

This paper is organized as follows: in Section \ref{sec:related}, we present
the archetypal analysis formulation; Section~\ref{sec:optim} is devoted to
optimization techniques; Section~\ref{sec:exp} presents successful applications
of archetypal analysis to computer vision tasks, and Section~\ref{sec:ccl}
concludes the paper.

\renewcommand{\thefootnote}{\arabic{footnote}}
\section{Formulation}\label{sec:related}
Let us consider a matrix $\X = [\x_1,\ldots,\x_n]$ in $\Real^{m \times n}$,
where each column $\x_i$ is a vector in $\Real^m$ representing some data point.
Archetypal analysis learns a factorial representation of~$\X$; it looks for
a set of~$p$ archetypes $\Z =[\z_1,\ldots,\z_p]$ in~$\Real^{m \times p}$ under
two geometrical constraints: each data vector $\x_i$ should be well
approximated by a convex combination of archetypes, and each
archetype~$\z_j$ should be a convex combination of data points~$\x_i$.
Therefore, given a set of archetypes~$\Z$, each vector
$\x_i$ should be close to a product $\Z \alphab_i$, where~$\alphab_i$
is a coefficient vector in the simplex $\Delta_p$:
\begin{equation}
   \Delta_p \defin \textstyle \left\{ \alphab \in \Real^p \st \alphab \geq 0
   ~~\text{and}~~ \sum_{j=1}^p \alphab[j] =1 \right\}. \label{eq:simplex}
\end{equation}
Similarly, for every archetype $\z_j$, there exists a vector $\betab_j$
in~$\Delta_n$ such that~$\z_j=\X\betab_j$, where $\Delta_n$ is defined
as in~(\ref{eq:simplex}) by replacing $p$ by~$n$.
Then, archetypal analysis is defined as a matrix factorization problem:
\begin{equation}
   \vspace*{-0.0cm}
   \min_{ \substack{\alphab_i \in \Delta_p ~\text{for}~ 1 \leq i \leq n\\
   \betab_j \in \Delta_n ~\text{for}~ 1 \leq j \leq p  }} \: \:
   \|\X-\X\B\A\|_{\text{F}}^2, \label{eq:archetypes}
   \vspace*{-0.1cm}
\end{equation}
where $\A = [\alphab_1, \ldots,\alphab_n]$,~$\B = [\betab_1,\ldots,\betab_p]$,
and $\|.\|_{\text{F}}$ denotes the Frobenius norm; the archetypes~$\Z$ are represented
by the product $\Z=\X\B$. Solving~(\ref{eq:archetypes}) is challenging since the
optimization problem is non-convex; this issue will be addressed in
Section~\ref{sec:optim}.  Interestingly, the formulation~(\ref{eq:archetypes})
is related to other approaches, which we briefly review here.

\vspace*{-0.3cm}
\paragraph{Non-negative matrix factorization (NMF)~\cite{Paa94}.} Assume that the
data~$\X$ is non-negative. NMF seeks for a factorization of $\X$ into two non-negative 
components:
\begin{equation}
   \vspace*{-0.0cm}
   \min_{\Z \in \Real^{m \times p}_+, \A \in \Real^{p \times n}_+} \|\X - \Z\A\|_{\text{F}}^2.
   \vspace*{-0.1cm}
\end{equation}
Similarly, the matrices $\Z$ and~$\A$ in archetypal analysis are also
non-negative when~$\X$ is itself non-negative. The difference between NMF and
archetypal analysis is that the latter involves simplicial constraints. 

\vspace*{-0.3cm}
\paragraph{Sparse Coding~\cite{Ols96}.}
Given a fixed set of archetypes~$\Z$ in~$\Real^{m \times p}$, each data
point~$\x_i$ is approximated by~$\Z\alphab_i$, under the constraint that
$\alphab_i$ is non-negative and sums to one. In other words, the~$\ell_1$-norm
of $\alphab_i$ is constrained to be one, which has a sparsity-inducing
effect~\cite{Bac12}. Thus, archetypal analysis produces sparse approximations
of the input data, and archetypes play the same role as the ``dictionary elements''
in the following sparse coding formulation of~\cite{Ols96}: 
\begin{equation}
   \vspace*{-0.0cm}
   \min_{\substack{\Z \in \Real^{m \times p}, \\ \A \in \Real^{p \times n}}} \frac{1}{2}\|\X - \Z\A\|_{\text{F}}^2 + \lambda\|\A\|_1  \st  \|\z_j\|_2 \leq 1 ~~\forall j. \label{eq:sparsecoding}
   \vspace*{-0.1cm}
\end{equation} Since the $\ell_1$-norm is related to the
simplicial constraints $\alphab_i \in \Delta_p$---the non-negativity
constraints being put aside---the main difference between sparse
coding and archetypal analysis is the fact that archetypes should be
convex combinations of the data points~$\X$.  As a result, the vectors
$\betab_j$ are constrained to be in the
simplex~$\Delta_n$, which encourages them to be sparse. Then, each
archetype~$\z_j$ becomes a linear combination of a few data points only,
which is useful for interpreting~$\z_j$.
Moreover, the non-zero entries in~$\betab_j$ indicate in which proportions
the input data points~$\x_i$ are related to each archetype~$\z_j$.

Another variant of sparse coding called ``local coordinate
coding''~\cite{YuK09} is also related to archetypal analysis.  In this variant,
dictionary elements are encouraged to be close to the data points that uses
them in their decompositions. Then, dictionary elements can be interpreted as
anchor points on a manifold representing the data distribution. 

\subsection{Robust Archetypal Analysis}
\label{sec:raa}
In some applications, it is desirable to automatically handle outliers---that
is, data points~$\x_i$ that significantly differ from the rest of the data. In order to make archetypal analysis robust, we
propose the following variant:
\begin{equation}
   \vspace*{-0.0cm}
   \min_{ \substack{\alphab_i \in \Delta_p ~\text{for}~ 1 \leq i \leq n\\
   \betab_j \in \Delta_n ~\text{for}~ 1 \leq j \leq p  }} \:\:
   \sum_{i=1}^n h\left(\|\x_i-\X\B\alphab_i\|_2\right), \label{eq:robust_archetypes}
   \vspace*{-0.1cm}
\end{equation}
where~$h: \Real \mapsto \Real$ is  the Huber loss function, which is often
used as a robust replacement of the squared loss in robust statistics~\cite{lange2000optimization}.
It is defined here for any scalar $u$ in~$\Real$ as
\begin{equation}
   \vspace*{-0.0cm}
   h(u) = \left\{ 
      \begin{array}{lr} \frac{u^2}{2\varepsilon} +\frac{\varepsilon}{2} & \text{if}~~  |u| \leq \varepsilon  \\ 
                      |u|  & \text{otherwise} \end{array} \right.,  \label{eq:huber}
   \vspace*{-0.1cm}
\end{equation}
and $\varepsilon$ is a positive constant. Whereas the cost associated to outliers in
the original formulation~(\ref{eq:archetypes}) can be large
since it grows quadratically, the Huber cost only grows linearly. In
Section~\ref{sec:optim}, we present an effective iterative reweighted least-square
strategy to deal with the Huber loss.

\section{Optimization}\label{sec:optim}
The formulation~(\ref{eq:archetypes}) is non-convex, but it is convex with
respect to one of the variables $\A$ or~$\B$ when the other one is fixed.  It
is thus natural to use a block-coordinate descent scheme, which is guaranteed
to asymptotically provide a stationary point of the problem~\cite{Ber99}. We present such 
a strategy in Algorithm~\ref{alg:archetypes}.
As noticed in~\cite{Cut94}, when fixing all variables but a
column~$\alphab_i$ of~$\A$ and minimizing with respect to~$\alphab_i$, the
problem to solve is a quadratic program (QP):
\begin{equation}
   \min_{\alphab_i \in \Delta_p} \|\x_i -\Z\alphab_i\|_2^2.\label{eq:updatealpha}
\end{equation}
These updates are carried out on Line~\ref{subalg:updatea} of Algorithm~\ref{alg:archetypes}.
Similarly, when fixing all variables but one column $\betab_j$ of~$\B$, we also
obtain a quadratic program:
\begin{equation}
   \min_{\betab_j \in \Delta_n} \|\X - \X\B_{\text{old}}\A + \X (\betab_{j,\text{old}} - \betab_j) \alphab^j\|_\text{F}^2, \label{eq:updatebeta}
\end{equation}
where $\betab_{j,\text{old}}$ is the current value of $\betab_j$ before the
update, and~$\alphab^j$ in~$\Real^{1 \times n}$ is the $j$-th row of~$\A$.
After a short calculation, problem~(\ref{eq:updatebeta}) can be equivalently rewritten
\begin{equation}
   \!\!  \min_{\betab_j \in \Delta_n} \left\| \frac{1}{\|\alphab^j\|_2^2}\left(\X \!-\! \X\B_{\text{old}}\A\right) \alphab^{j \top}  \!+\! \X \betab_{j,\text{old}} \!-\! \X\betab_j\right\|_2^2, \label{eq:updatebeta2}
\end{equation}
which has a similar form as~(\ref{eq:updatealpha}).
This update is carried out on Line~\ref{subalg:updateb} of
Algorithm~\ref{alg:archetypes}.
Lines~\ref{subalg:updatez} and~\ref{subalg:updateR} respectively update the archetypes and
the residual~$\R=\X-\X\B\A$.
Thus, Algorithm~\ref{alg:archetypes} is a cyclic
block-coordinate algorithm, which is guaranteed to converge to a stationary point
of the optimization problem~(\ref{eq:archetypes}), see, e.g.,~\cite{Ber99}.
\begin{algorithm}[tb]
    \caption{Archetypal Analysis}
    \label{alg:archetypes}
    \begin{algorithmic}[1]
    \STATE {\bfseries Input:} Data $\X$ in~$\Real^{m \times n}$; $p$ (number of archetypes); \\ $T$ (number of iterations);
    \STATE Initialize $\Z$ in~$\Real^{m \times p}$ with random columns from~$\X$; 
    \STATE Initialize $\B$ such that $\Z=\X\B$;
    \FOR {$t=1\ldots,T$}
    \FOR  {$i=1\ldots,n$}
    \STATE  \label{subalg:updatea}
$       \alphab_i \in \argmin_{\alphab \in \Delta_p} \|\x_i -\Z\alphab\|_2^2$;
    \ENDFOR
    \STATE $\R \leftarrow \X-\Z\A$;
    \FOR  {$j=1\ldots,p$}
    \STATE $\betab_j \in \argmin_{\betab \in \Delta_n} \left\| \frac{1}{\|\alphab^j\|_2^2} \R \alphab^{j \top} +\z_j -\X\betab\right\|_2^2$; \label{subalg:updateb}
    \STATE $\R \leftarrow \R + (\z_j - \X\betab_j)\alphab^j$; \label{subalg:updateR}
       \STATE $\z_j \leftarrow \X\betab_j$;\label{subalg:updatez}
    \ENDFOR
    \ENDFOR
    \STATE \textbf{Return} $\A$, $\B$ (decomposition matrices).
 \end{algorithmic}
 \end{algorithm}
The main difficulty to implement such a strategy is to find
a way to efficiently solve quadratic programs with simplex constraints such
as~(\ref{eq:updatealpha}) and~(\ref{eq:updatebeta2}). We discuss this issue
in the next section.

\subsection{Efficient Solver for QP with Simplex Constraint}
Both problems~(\ref{eq:updatealpha}) and~(\ref{eq:updatebeta2}) have the same
form, and thus we will focus on finding an algorithm for
solving:
\begin{equation}
   \min_{\alphab \in \Delta_p} \left[f(\alphab) \defin  \|\x -\Z\alphab\|_2^2\right], 
\end{equation}
which is a smooth (least-squares) optimization problem with a simplicial
constraint.  Even though generic QP solvers could be used, significantly faster
convergence can be obtained by designing a dedicated algorithm that can
leverage the underlying ``sparsity'' of the solution~\cite{Bac12}.

We propose to use an active-set algorithm~\cite{Noc06} that can benefit
from the solution sparsity, when carefully implemented. 
Indeed, at the optimum, most often only a small
subset~$\AA$ of the variables will be non-zero. 
Active-set algorithms~\cite{Noc06} can be seen as an aggressive strategy that
can leverage this property. 
Given a current estimate $\alphab$ in~$\Delta_p$ at some iteration,
they define a subset~$\AA = \{ j \st \alphab[j] > 0\}$, and
find a direction $\q$ in~$\Real^{p}$ by solving the reduced problem
\begin{equation}
   \min_{\q \in \Real^p} \|\x -\Z(\alphab + \q)\|_2^2 \st \sum_{j =1}^{p} \q[j] = 0 ~\text{and}~ \q_{\AA^C}=0,
   \label{eq:eqQP}
\end{equation}
where $\AA^C$ denotes the complement of $\AA$ in the index set $\{1\ldots,p\}$.
Then, a new estimate $\alphab' \!=\! \alphab\!+\!\gamma\q$ is obtained by moving~$\alphab$ onto the direction~$\q$---that is, choosing $\gamma$ in~$[0,1]$, such that~$\alphab'$ remains
in~$\Delta_p$.  The algorithm modifies the set~$\AA$ until the algorithm
finds an optimal solution in~$\Delta_p$. This strategy
is detailed in Algorithm~\ref{alg:activeSet}.
 \begin{algorithm}[bt]
   \caption{Active-Set Method}
   \label{alg:activeSet}
   \begin{algorithmic}[1]
   \STATE {\bfseries Input:} matrix $\Z \in \mathbb{R}^{m \times p}$, 
vector $\x \in \mathbb{R}^m$;
   \STATE Initialize $\alphab_0 \in \Delta_p$ with a feasible starting point;
   \STATE Define $\AA_0 \leftarrow \{ j \st \alphab_0[j] > 0\}$;
   \FOR {$k = 0, 1, 2...$}
   \STATE Solve~(\ref{eq:eqQP}) with $\AA_k$ 
to find a step $\q_k$;
   \IF {$\q_k$ is $0$}
\STATE Compute $\nabla f(\alphab_k)\!=\!-2\Z^\top(\x\!-\!\Z\alphab_k)$; 
   \IF { $\nabla f(\alphab_k)[j] > 0$ for all $j \notin \AA_k$}
     \STATE \textbf{Return} $\alphab^* = \alphab_k$ (solution is optimal).
     \ELSE 
  \STATE $j^\star \leftarrow \min_{j \notin \AA_k} \nabla f(\alphab_k)[j]$;
     \STATE $\AA_{k+1}  \leftarrow \AA_{k} \cup \{j^\star\}$;
     \ENDIF
   \ELSE
   \STATE $\gamma_k \leftarrow \max_{\gamma \in [0,1]}\left[ \gamma \st {\alphab_k+\gamma\q_k \in \Delta_p}\right]$;
   \IF {$\gamma_k < 1$} 
   \STATE Find $j^\star$ such that $\alphab_k[j]+\gamma_k\q_k[j]=0$;
   \STATE $\AA_{k+1} \leftarrow \AA_{k} \setminus \{ j^\star \}$; 
   \ELSE 
   \STATE $\AA_{k+1} \leftarrow \AA_{k}$;
\ENDIF
\STATE $\alphab_{k+1} \leftarrow \alphab_k + \gamma_k \q_k$;
\ENDIF
   \ENDFOR
\end{algorithmic}
\end{algorithm}
Open-source active-set solvers for generic QP exist, \eg, quadprog in Matlab,
but we have found them too slow for our purpose. Instead, a dedicated
implementation has proven to be much more efficient. More precisely, we use
some tricks inspired from the Lasso solver of the SPAMS
toolbox~\cite{mairal2010online}: (i) initialize~$\AA$ with a single variable;
(ii) update at each iteration the quantity~$(\Z_\AA^\top\Z_\AA)^{-1}$ by using Woodbury formula;
(iii) implicitly working with the matrix $\Q=\Z^\top\Z$ without computing it when updating~$\betab$.

As a resutl, each iteration of the active-set algorithm has a computational complexity of 
${O}(mp+{a}^2)$ operations where $a$ is the size of the set~$\AA$ at that 
iteration. Like the simplex algorithm for solving linear programs~\cite{Noc06}, the
maximum number of iterations of the active-set algorithm can be exponential in
theory, even though it is much smaller than $\min(m,p)$ in practice.
Other approaches than the active-set algorithm could be considered, such as the
fast iterative shrinkage-thresholding algorithm (FISTA) and the penalty
approach of~\cite{Cut94}. However, in our experiments, we have observed
significantly better performance of the active-set algorithm, both in terms of
speed and accuracy. 

\subsection{Optimization for Robust Archetypal Analysis}
To deal with the Huber loss, we use the
following variational representation of the Huber loss~(see, \cite{lange2000optimization}):
\begin{equation}
   \vspace*{-0.1cm}
   h(u) = \frac{1}{2} \min_{w \geq \varepsilon} \left[\frac{u^2}{w} + w\right]. \label{eq:huber_var}
\end{equation} 
which is equivalent to~(\ref{eq:huber}). 
Then, robust archetypal analysis from Eq.~(\ref{eq:robust_archetypes}) can be
reformulated as 
\begin{equation}
   \vspace*{-0.1cm}
   \min_{ \substack{\alphab_i \in \Delta_p ~\text{for}~ 1 \leq i \leq n\\
   \betab_j \in \Delta_n ~\text{for}~ 1 \leq j \leq p  \\ w_i \geq \varepsilon ~\text{for}~ 1 \leq i \leq n}} \: \:
   \frac{1}{2}\sum_{i=1}^n \frac{1}{w_i}\|\x_i-\X\B\alphab_i\|_2^2 +  w_i. \label{eq:robust_archetypes2}
\end{equation}
We have introduced here one weight $w_i$ per data point. Typically, $(1/w_i)$
becomes small for outliers, reducing their importance in the objective
function. Denoting by~$\w=[w_1,\ldots,w_n]$ the weight vector in~$\Real^n$,
the formulation~(\ref{eq:robust_archetypes2}) has the
following properties: (i) when fixing all
variables $\A,\B,\w$ but one vector~$\alphab_i$, and optimizing with respect to
~$\alphab_i$ we still obtain a quadratic program with simplicial constraints;
(ii) the same is true for the vectors~$\betab_j$; (iii) when fixing~$\A$
and~$\B$, optimizing with respect to $\w$ has a closed form solution.
It is thus natural to use the block-coordinate descent scheme, which
is presented in Algorithm~\ref{alg:robust_archetypes}, and which 
is guaranteed to converge to a stationary point. 

\begin{algorithm}[tb]
    \caption{Robust Archetypal Analysis}
    \label{alg:robust_archetypes}
    \begin{algorithmic}[1]
    \STATE {\bfseries Input:} Data $\X$ in~$\Real^{m \times n}$; $p$ (number of archetypes); \\ $T$ (number of iterations);
    \STATE Initialize $\Z$ in~$\Real^{m \times p}$ with random columns from~$\X$; 
    \STATE Initialize $\B$ such that $\Z=\X\B$;
    \STATE Initialize $\w$ in $\Real^n$ with $w_i=1$ for all $1\leq i \leq n$;
    \FOR {$t=1\ldots,T$}
    \FOR  {$i=1\ldots,n$}
    \STATE  \label{subalg:updatea2}
$       \alphab_i \in \argmin_{\alphab \in \Delta_p} \|\x_i -\Z\alphab\|_2^2;$
    \STATE $w_i \leftarrow \max(\|\x_i -\Z\alphab\|_2,\varepsilon)$;
    \ENDFOR
    \STATE $\Gammab \leftarrow \diag(\w)^{-1}$ (scaling matrix);
    \STATE $\R \leftarrow \X-\Z\A$;
    \FOR  {$j=1\ldots,p$}
    \STATE $\betab_j \in \argmin_{\betab \in \Delta_n} \left\| \frac{ \R \Gammab \alphab^{j \top}}{\alphab^j \Gammab \alphab^{j \top}}  +\z_j -\X\betab\right\|_2^2$; \label{subalg:updateb2}
    \STATE $\R \leftarrow \R + (\z_j - \X\betab_j)\alphab^j$; \label{subalg:updateR2}
       \STATE $\z_j \leftarrow \X\betab_j$;\label{subalg:updatez2}
    \ENDFOR
    \ENDFOR
    \STATE \textbf{Return} $\A$, $\B$ (decomposition matrices).
 \end{algorithmic}
\end{algorithm}

The differences between Algorithms~\ref{alg:archetypes}
and~\ref{alg:robust_archetypes} are the following: each time a
vector~$\alphab_i$ is updated, the corresponding weight $w_i$
is updated as well, where $w_i$ is the solution of Eq.~(\ref{eq:huber_var})
with $u= \|\x_i -\Z\alphab\|_2$. 
The solution is actually $w_i = \max(\|\x_i -\Z\alphab\|_2, \varepsilon)$.
Then, the update of the vectors~$\betab_j$ is slightly more involved.
Updating~$\betab_j$ yields the following optimization problem:
\begin{equation}
   \min_{\betab_j \in \Delta_n} \left\|\left(\X - \X\B_{\text{old}}\A + \X (\betab_{j,\text{old}} - \betab_j) \alphab^j\right) \Gammab^{1/2}\right\|_\text{F}^2,  \label{eq:robust_update}
\end{equation}
where the diagonal matrix~$\Gammab$ in~$\Real^{n \times n}$ carries the inverse
of the weights~$\w$ on its diagonal, thus rescaling the residual of each data
point~$\x_i$ by $1/w_i$ as in~(\ref{eq:robust_archetypes2}).
Then, it is possible to show that~(\ref{eq:robust_update}) is 
equivalent to 
\begin{displaymath}
   \min_{\betab_j \in \Delta_n} \left\|\frac{1}{\alphab^{j T}\Gammab \alphab^j}\left(\X \!-\! \X\B_{\text{old}}\A\right)\Gammab\alphab^{j \top} + \X \betab_{j,\text{old}} \!- \!\X\betab_j\right\|_2^2,
\end{displaymath}
which is carried out on Line~\ref{subalg:updateb2} of Algorithm~\ref{alg:robust_archetypes}.

\section{Experiments}\label{sec:exp}
We now study the efficiency of archetypal analysis in various applications. Our
implementation is coded in C++ and interfaced with R, Python, and Matlab. It has
been included in the toolbox SPAMS v2.5~\cite{mairal2010online}. The number of
iterations for archetypal analysis was set to $T=100$, which leads to a good
performance in all experiments.

\subsection{Comparison with Other Implementations}
To the best of our knowledge, two software packages implementing archetypal analysis are publicly available:\\
\hspace*{0.2cm}~$\bullet$ The Python Matrix Factorization toolbox
(PyMF)\footnote{\url{http://code.google.com/p/pymf/}.} is an open-source library
that tackles several matrix factorization problems including archetypal
analysis. It performs an alternate minimization scheme between the $\alphab_i$'s
and~$\betab_j$'s, but relies on a generic QP solver from CVX.\footnote{\url{http://cvxr.com/cvx/}.} \\
\hspace*{0.2cm}~$\bullet$ The R package
\textsf{archetypes}\footnote{\url{http://archetypes.r-forge.r-project.org/}.} is
the reference implementation of archetypal analysis for R, which
is one of the most widely used high-level programming language in statistics.
Note that the algorithm implemented in this package deviates from the original
archetypal analysis described in~\cite{Cut94}.

\begin{figure}[hbtp]
   \includegraphics[width=0.49\linewidth,trim=20 0 80 0]{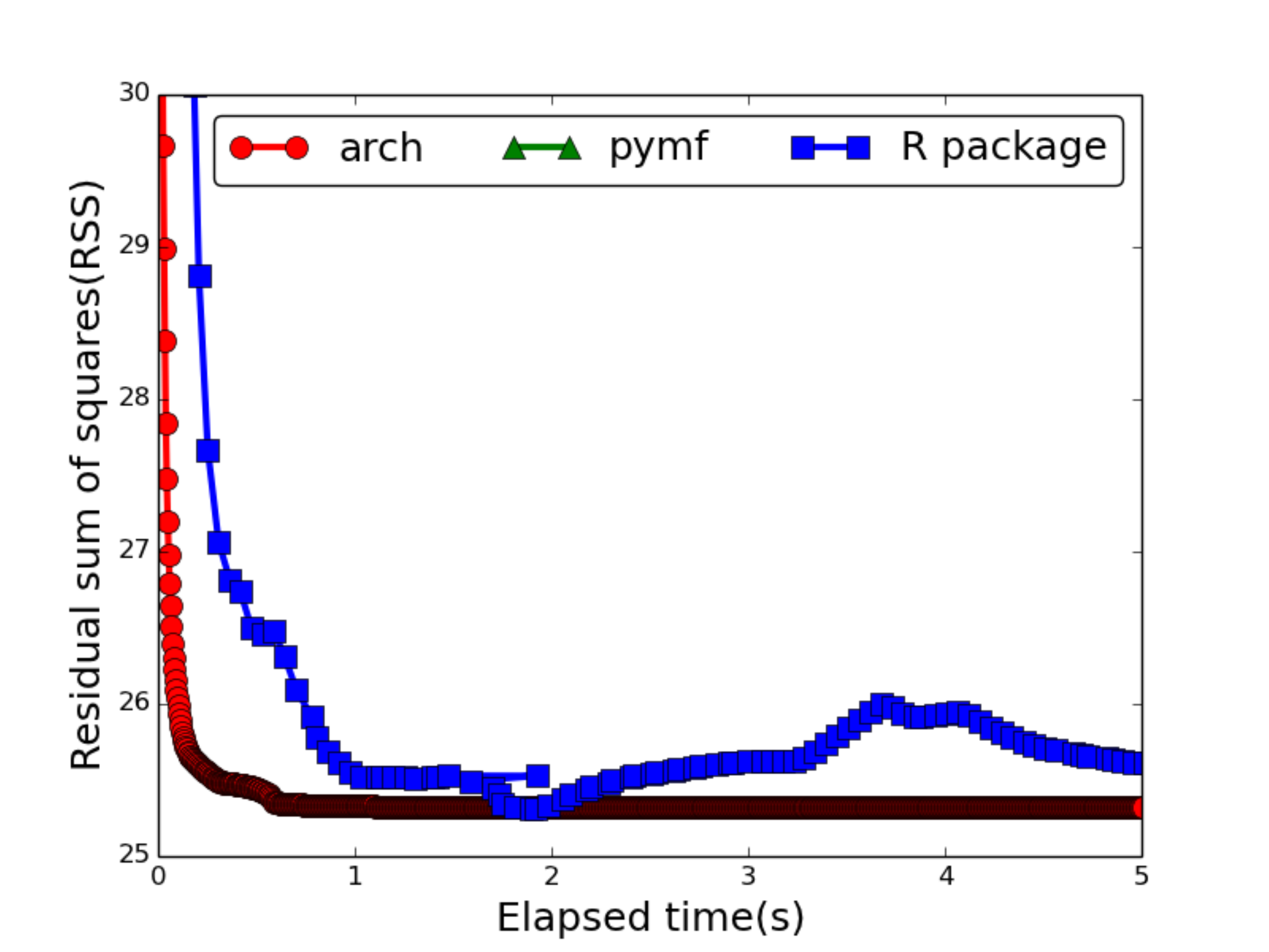} \hfill
   \includegraphics[width=0.49\linewidth,trim=20 0 80 0]{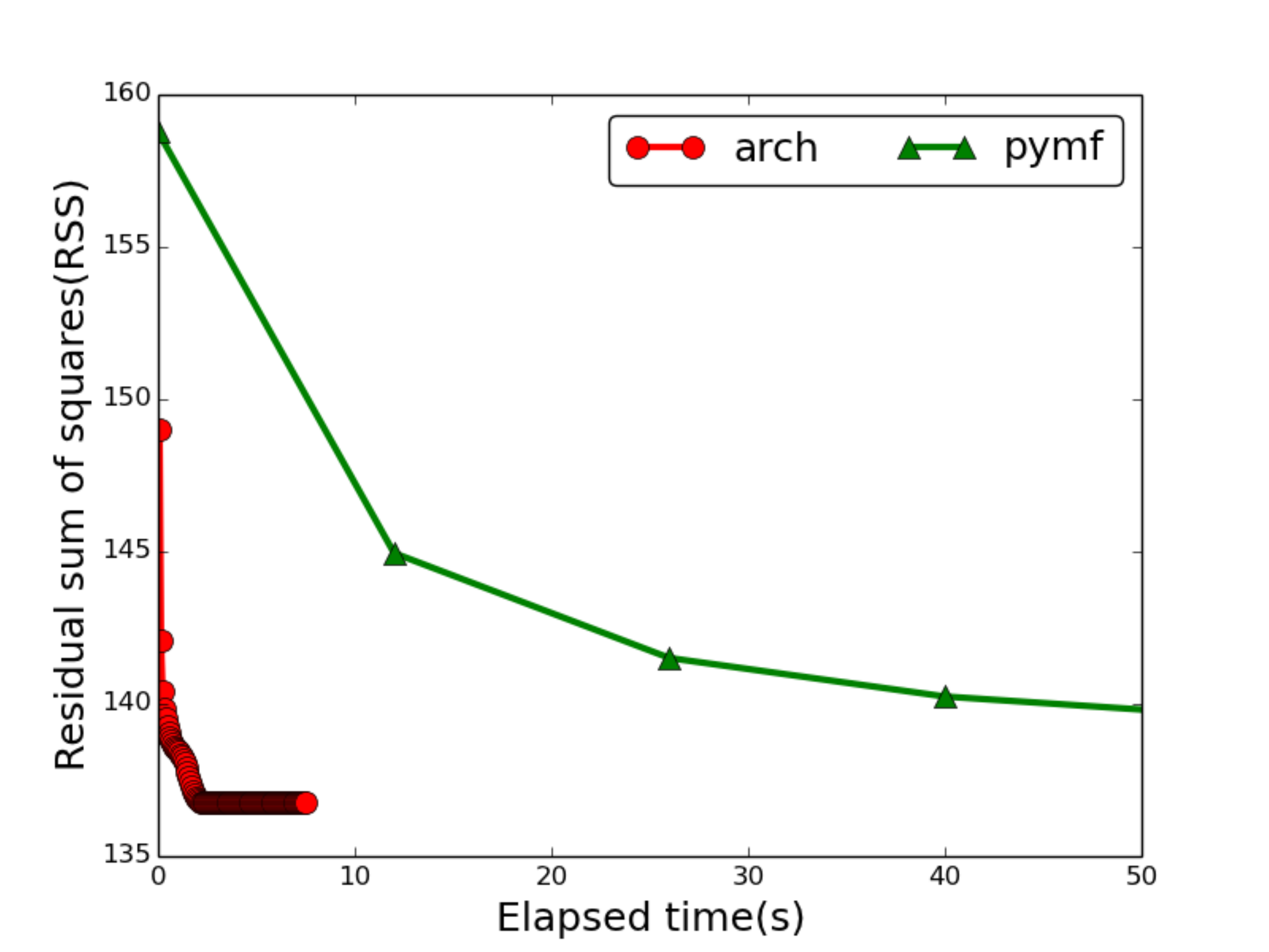} 
   \caption{Experimental comparison with other implementations.
      Left: value of the objective function vs computational 
      time for a dataset with $m\!=\!10$, $n\!=\!507$ and $p\!=\!5$ archetypes. 
      Our method is denoted by~\textsf{arch}.
      PyMF was too slow to
      appear in the graph and the R-package exhibits a non-converging behavior.
      Right: same experiment with $n\!=\!600$ images from 
      MNIST~\cite{lecun1998gradient}, of size $m\!=\!784$, with $p\!=\!10$ archetypes.
      The R package diverged while PyMF was between $100$ and $1000$ times slower
   than our approach.}
   \label{fig:speed}
\end{figure}

We originally intended to try all methods on matrices~$\X$ in $\Real^{m
\times n}$ with different sizes $m$ and $n$ and different numbers $p$ of
archetypes. Unfortunately, the above software packages suffer from severe limitations and
we were only able to try them on small datasets.  We report such a
comparison in Figure~\ref{fig:speed}, where the computational times are
measured on a single core of an Intel Xeon CPU E5-1620.  We only report
results for the R package on the smallest dataset since it diverged on larger ones,
while PyMF was several orders of magnitudes slower than our implementation.
We also conducted an experiment following the optimization scheme of
PyMF but replacing the QP solver by other alternatives such as Mosek
or quadprog, and obtained similar conclusions.

\begin{figure}[hbtp]
   \includegraphics[width=0.49\linewidth,trim=10 0 70 0]{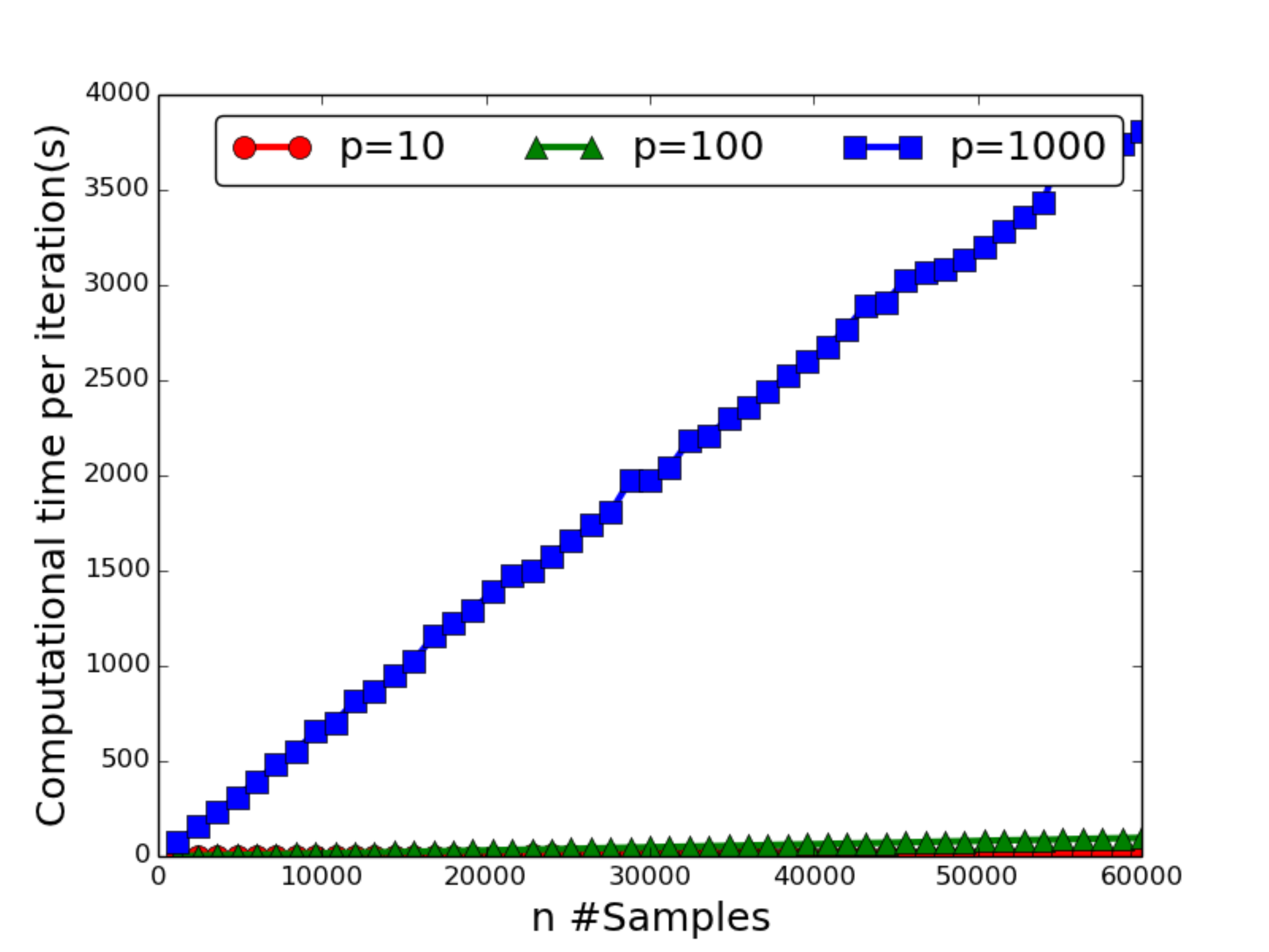} \hfill
   \includegraphics[width=0.49\linewidth,trim=10 0 70 0]{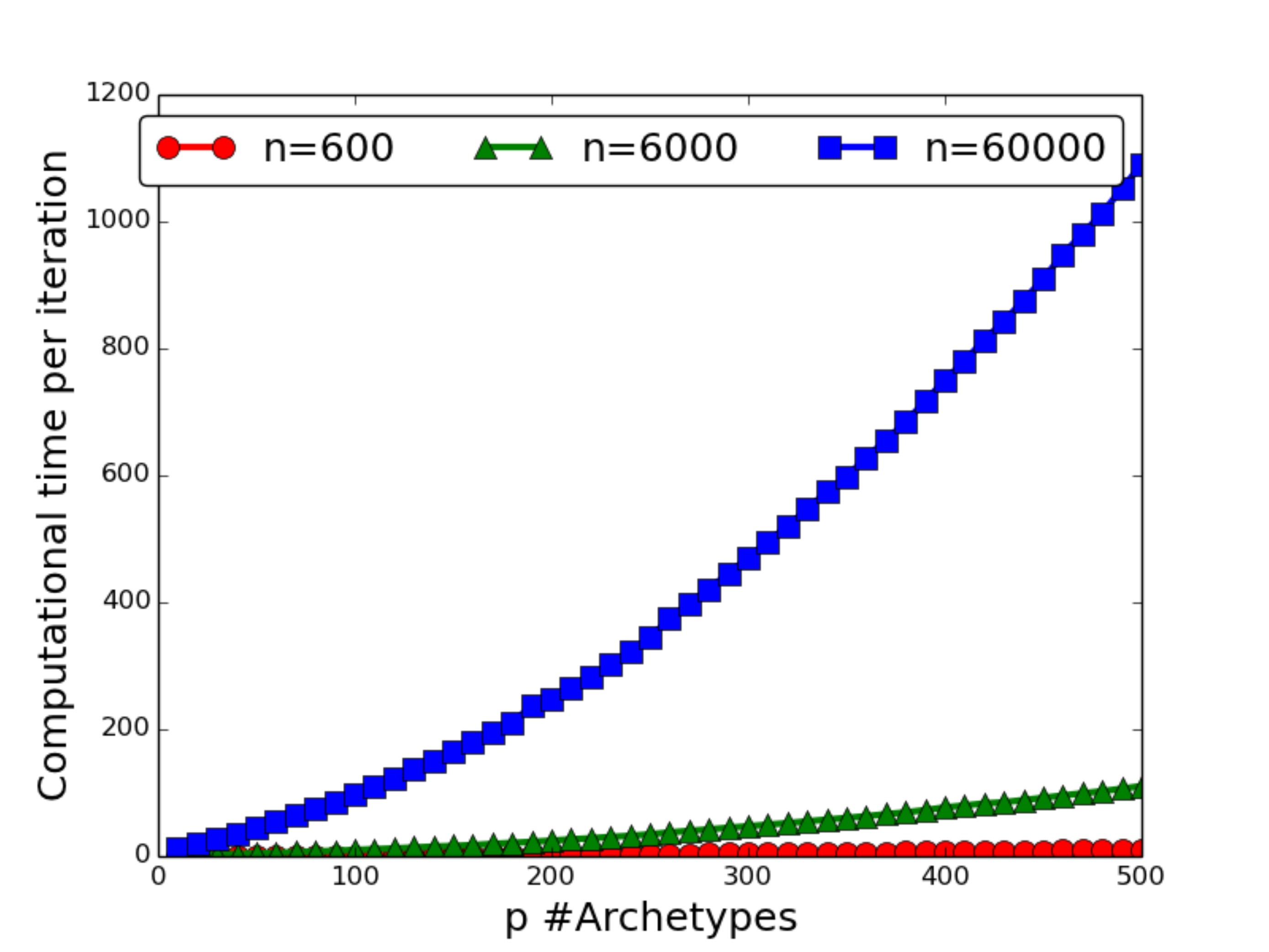} 
   \caption{Scalability Study. Left: the computational time per iteration when
      varying the sample size~$n$ for different numbers of archetypes~$p$. The
      complexity of our implementation is empirically linear in~$n$.
      Right: the same experiment when varying~$p$ and with fixed sample sizes $n$.
   The complexity is more than linear in~$p$.} \label{fig:speed2} 
\end{figure}

Then, we study the scalability of our implementation in regimes where
the R package and PyMF are unusable. We report in Figure~\ref{fig:speed2} the computational cost per
iteration of our method when varying~$n$ or~$p$ on the MNIST
dataset~\cite{lecun1998gradient}, where $m=784$. We observe that the empirical
complexity is approximately linear in~$n$, allowing us to potentially learn
large datasets with more than $100\,000$ samples, and above linear in~$p$,
which is thus the main limitation of our approach.  However, such a limitation
is also shared by classical sparse coding techniques, where the empirical
complexity regarding~$p$ is also greater than $O(p)$~\cite{Bac12}.

\subsection{Codebook Learning with Archetypal Analysis}\label{subsec:codebook}
Many computer vision approaches have represented so far images
under the form of a ``bag of visual words'', or by using some variants of it.
In a nutshell, each local patch (small regions of typically $16 \times 16$
pixels) of an image is encoded by a descriptor which is invariant to small
deformations, such as SIFT~\cite{Laz06}. Then, an unsupervised learning technique is used for
defining a codebook of visual patterns called ``visual words''. The image is
finally described by computing a histogram of word occurrences, yielding a powerful
representation for discriminative tasks~\cite{Csu04,Laz06}. 

More precisely, typical methods for learning the codebook are $K$-means and
sparse coding~\cite{Laz06,Yan09}.  SIFT descriptors are sparsely encoded
in~\cite{Yan09} by using the formulation~(\ref{eq:sparsecoding}), and the image
representation is obtained by ``max-pooling'' the sparse codes, as explained
in~\cite{Yan09}.   Spatial pyramid matching (SPM) \cite{Laz06} is also used,
which includes some spatial information, yielding better accuracy than simple
bags of words on many benchmark datasets.  Ultimately, the classification task
is performed with a support vector machine (SVM) with a linear kernel.

It is thus natural to wonder whether a similar performance could be achieved by
using archetypal analysis instead of sparse coding.  We thus conducted an image
classification experiment by using the software package of~\cite{Yan09}, and
simply replacing the sparse coding component with our implementation of
archetypal analysis. We use as many archetypes as dictionary elements
in~\cite{Yan09}---that is, $p=1\,024$, and $n=200\,000$ training samples, and
we call the resulting method ``archetypal-SPM''. We use the same datasets
as~\cite{Yan09}---that is, Caltech-101 \cite{Fei07} and 15 Scenes
Categorization \cite{Fei05, Laz06}. The purpose of this experiment is to
demonstrate that archetypal analysis is able to learn a codebook that is as good as
sparse coding and better than K-means.  Thus, only results of similar methods
are represented here such as~\cite{Laz06,Yan09}. The state of the art on these
data sets may be slightly better nowadays, but involves a different recognition
pipeline. We report the results in Tables~\ref{table:caltech}
and~\ref{table:scenes}, where archetypal analysis seems to perform as well
as sparse coding.  Note that KMeans-SPM-$\chi^2$ uses a $\chi^2$-kernel for
the SVM~\cite{Laz06}.

\begin{table}[hbtp]
   \begin{center}
      \begin{tabular}{|lcc|}
         \hline
         Algorithms          & 15 training      & 30 training \\
         \hline\hline
         KMeans-SPM-$\chi^2$~\cite{Laz06}   & 56.44 $\pm$ 0.78    & 63.99 $\pm$ 0.88 \\
                   KMeans-SPM~\cite{Yan09}  & 53.23 $\pm$ 0.65 & 58.81 $\pm$ 1.51 \\ 
                        SC-SPM~\cite{Yan09} & 67.00 $\pm$ 0.45 & 73.20 $\pm$ 0.54 \\
                            archetypal-SPM  & {64.96 $\pm$ 1.04} & {72.00 $\pm$ 0.88} \\
         \hline
      \end{tabular}
   \end{center}
   \vspace*{-0.4cm}
   \caption{Classification accuracy (\%) on Caltech-101 dataset. Following the same experiment in \cite{Yan09}, 15 or 30 images per class are randomly chosen for training and the rest for testing. The standard deviation is obtained with 10 randomized experiments. }
   \label{table:caltech}
\end{table}
\begin{table}[hbtp]
   \begin{center}
      \begin{tabular}{|lc|}
         \hline
         Algorithms          & Classification Accuracy \\
         \hline\hline
         KMeans-SPM-$\chi^2$~\cite{Laz06}   & 81.40 $\pm$ 0.50 \\
                    KMeans-SPM~\cite{Yan09} & 65.32 $\pm$ 1.02 \\
                        SC-SPM~\cite{Yan09} & 80.28 $\pm$ 0.93 \\
                       archetypal-SPM       & {81.57 $\pm$ 0.81}  \\
         \hline
      \end{tabular}
   \end{center}
   \vspace*{-0.4cm}
   \caption{Classification accuracy (\%) on Scene-15 dataset. Following the same experiment in \cite{Yan09}, 100 images are randomly chosen for training and the rest for testing. The standard deviation is obtained with 10 randomized experiments.}
   \label{table:scenes}
\end{table}

\subsection{Digit Classification with Archetypal Analysis}
\begin{figure}[hbtp]
   \includegraphics[width=0.49\linewidth,trim=30 0 36 0]{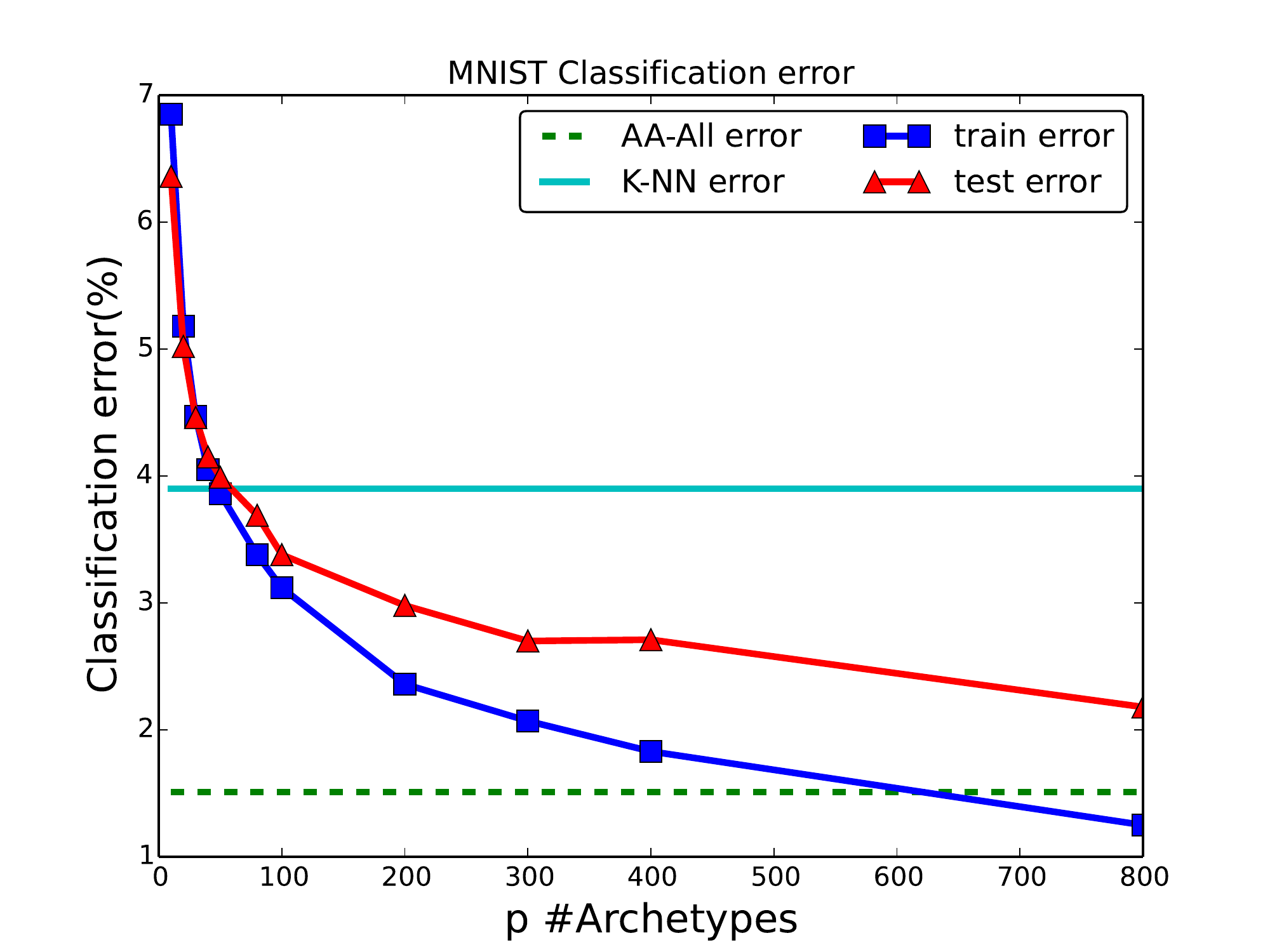} \hfill
   \includegraphics[width=0.49\linewidth,trim=30 0 36 0]{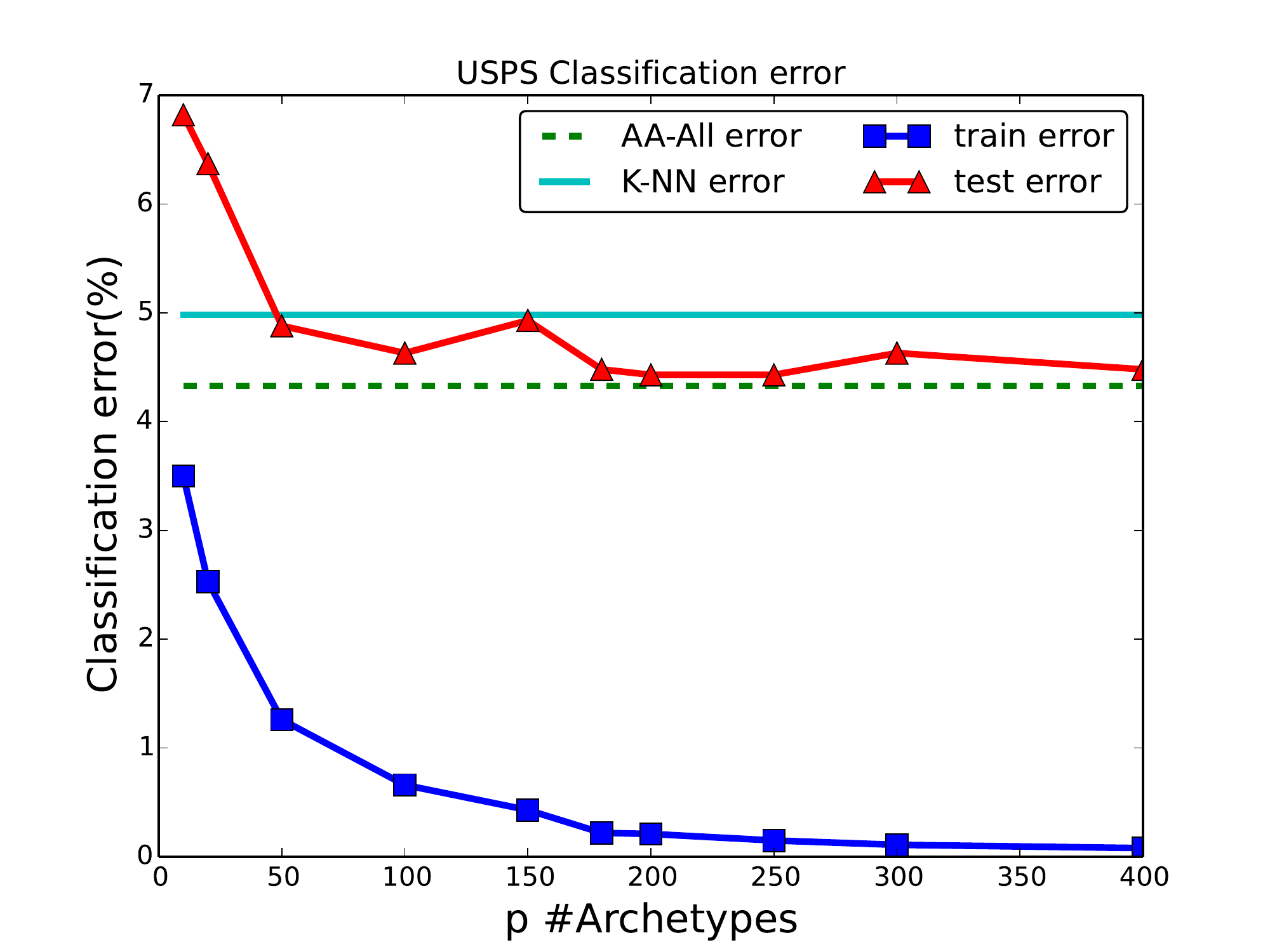} 
   \caption{Train and Test error on MNIST (left) and USPS (right) when varying
      the number~$p$ of archetypes. All-AA means that all data points are used as
   archetypes.} \label{fig:mnist} 
\end{figure}

Even though sparse coding is an unsupervised learning technique, it has been
used directly for classification tasks~\cite{Mai12,Ram10,Wri09}. 
For digit recognition, it has been observed in~\cite{Ram10} that simple
classification rules based on sparse coding yield impressive results on
classical datasets such as MNIST~\cite{lecun1998gradient} and USPS.
Suppose that one has learned on training data a dictionary~$\Z_k$
in~$\Real^{m \times p}$ for every digit class $k=0,\ldots,9$ by using~(\ref{eq:sparsecoding}), and that a new test digit~$\x$ in~$\Real^m$ is
observed. Then, $\x$ can be classified by finding the class~$k^\star$ that best
represents it with~$\Z_k$:
\begin{equation}
   k^\star = \argmin_{k \in \{0,\ldots,9\}} \left[\min_{\alphab \in \Real^p} \frac{1}{2}\|\x-\Z_k\alphab\|_2^2 + \lambda\|\alphab\|_1\right], \label{eq:class_sc}
\end{equation}
where $\lambda$ is set to $0.1$ in~\cite{Ram10} and the vectors~$\x$ are
normalized. Since we want to compare archetypal analysis with sparse coding, it is thus natural to also consider the
corresponding ``archetype'' classification rule:
\begin{equation}
   k^\star = \argmin_{k \in \{0,\ldots,9\}} \left[\min_{\alphab \in \Delta_p} \|\x-\Z_k\alphab\|_2^2\right], \label{eq:class_aa}
\end{equation}
where the $\Z_k$ are archetypes learned for every digit class.  Note that when
archetypes are made of all available training data, the convex dual
of~(\ref{eq:class_aa}) is equivalent to the nearest convex hull classifier
of~\cite{Nal06}.  We report the results when all the training data is used as
archetypes in Table~\ref{table:mnist}, and when varying
the number of archetypes per class in Figure~\ref{fig:mnist}. We include in
this comparison the performance of SVM with a Gaussian kernel, and the
$K$-nearest neighbor classifier (K-NN).  Even though the state of the art on MNIST
achieves less than $1\%$ test error~\cite{Mai12,Ran07}, the results reported
in Table~\ref{table:mnist} are remarkable for several reasons: (i) the method
AA-All has no hyper-parameter and performs almost as well as sparse coding,
which require choosing~$\lambda$; (ii) AA-All and SC-All significantly
outperform K-NN and perform similarly as a non-linear SVM, even though they
use a simple Euclidean norm for comparing two digits; (iii) none of the 
methods in Table~\ref{table:mnist} exploit the fact that the $\x_i$'s are in
fact images, unlike more sophisticated techniques such as convolutional neural
networks~\cite{Ran07}. 
In Figure~\ref{fig:mnist}, neither SC nor AA are helpful for prediction, but
archetypal analysis can be useful for reducing the computational cost at test
time. The choice of dictionary size $K$ should be driven by this trade-off.
For example, on USPS, using $200$ archetypes per class yields similar results
as AA-All. 

\begin{table}[hbtp]
   \begin{center}
      \begin{tabular}{|lcccc|}
         \hline
         Dataset  & AA-All & SC-All & SVM & K-NN \\
         \hline\hline
         MNIST & 1.51 & 1.35 & 1.4 & 3.9 \\
          USPS & 4.33  & 4.14 & 4.2 & 4.93 \\
         \hline
      \end{tabular}
   \end{center}
   \vspace*{-0.4cm}
   \caption{Classification error rates (\%) on the test set for the MNIST and USPS
      datasets. AA-All and SC-All respectively mean that all data points are used as
   archetypes and  dictionary elements. SVM uses a Gaussian kernel.}
   \label{table:mnist}
\end{table}

\subsection{Archetyping Flickr Requests}
Data visualization
has now become an important topic, especially regarding image databases from the
Internet~\cite{doersch2012makes}, or videos~\cite{elhamifar2012see}.  We focus
in this section on public images downloaded from the Flickr website, and
present a methodology for visualizing the content of different requests
using \emph{robust archetypal analysis} presented in Section~\ref{sec:raa}. 

For example, we present in this section a way to visualize the request
``Paris'' when downloading $36\,600$ images uploaded in $2012$ and $2013$, and
sorted by ``relevance'' according to Flickr. We first compute dense SIFT
descriptors~\cite{Laz06} for all images, and represent
each image by using a Fisher vector~\cite{Per10}, which have shown good
discriminative power in classification tasks. Then, we learn $p=256$
archetypes.  Interestingly, we observe that the Flickr request has a large
number of  outliers, meaning that some images tagged as ``Paris'' are actually
unrelated to the city. Thus, we choose to use the robust version of archetypal
analysis in order to reduce the influence of such outliers. We use similar heuristics for choosing $\epsilon$ 
as in the robust statistics literature, resulting in $\epsilon=0.01$ for data points that are $\ell_2$-normalized.

\begin{figure*}[hbtp]
   \subfloat[Archetypes representing (expected) landmarks.]{\label{subfig:paris1}
      \begin{minipage}{0.33\textwidth}
         \includegraphics[width=\textwidth]{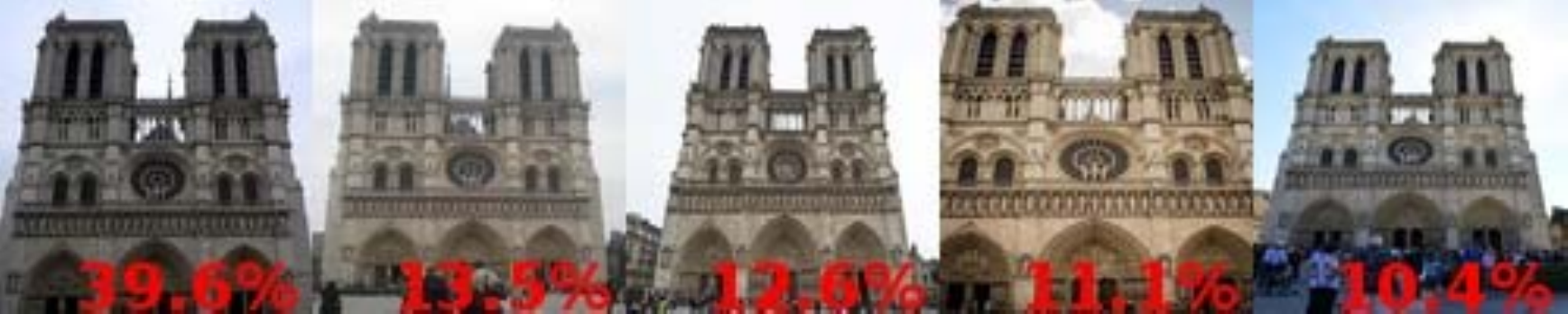} \newline
         \includegraphics[width=\textwidth]{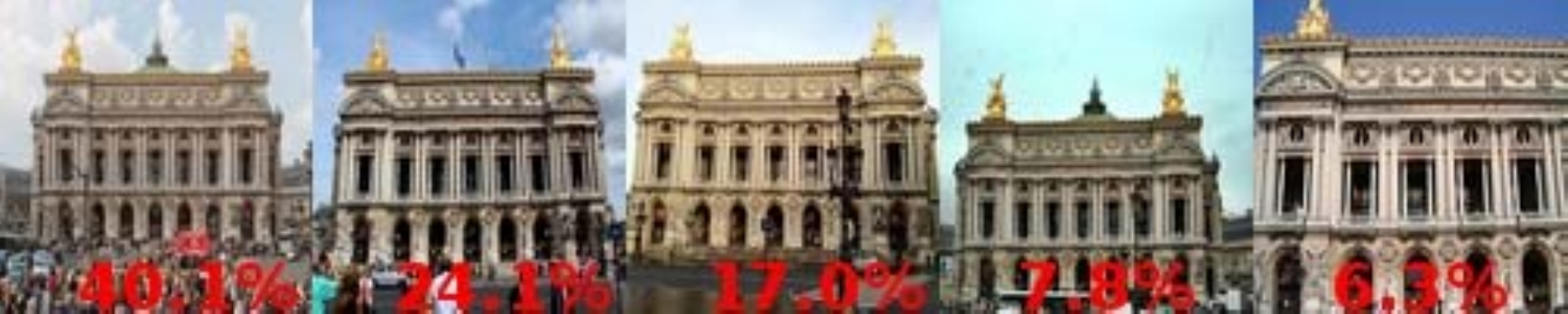} \\
         \includegraphics[width=\textwidth]{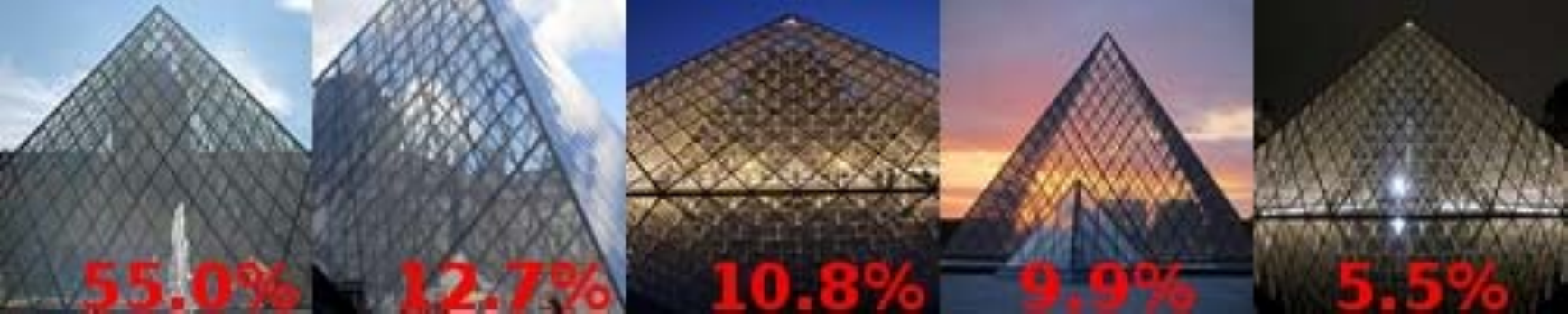} \\
         \includegraphics[width=\textwidth]{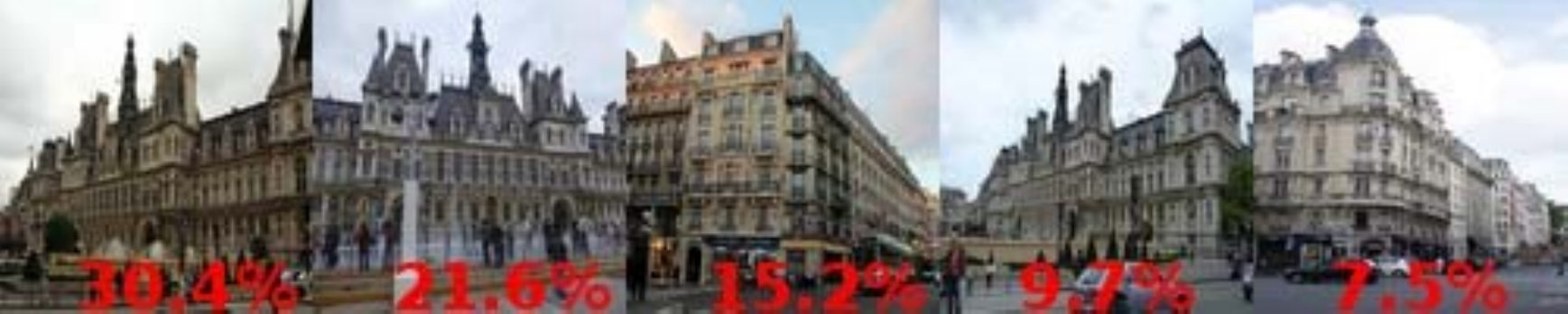} \\
         \includegraphics[width=\textwidth]{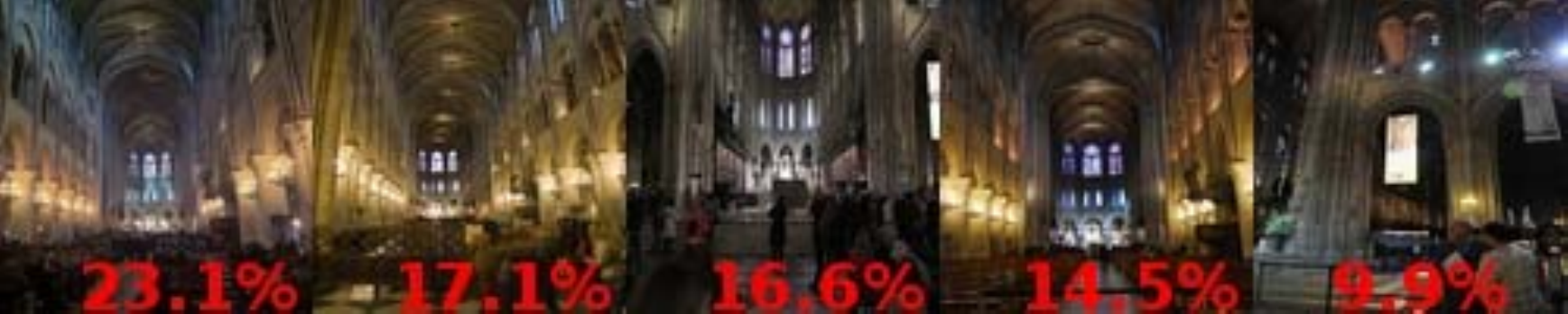} \\
         \includegraphics[width=\textwidth]{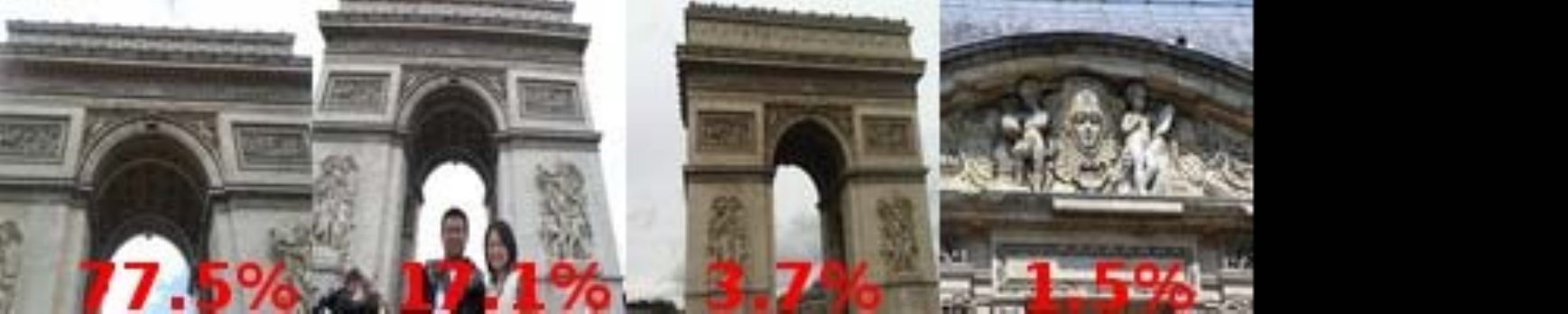} \\
         \includegraphics[width=\textwidth]{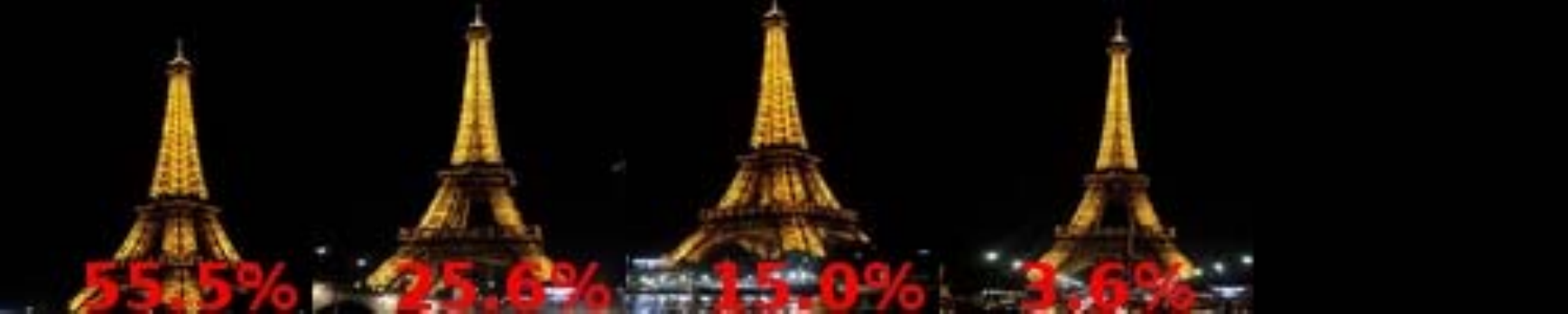} 
      \end{minipage}
   }
   \subfloat[Less expected archetypes.]{\label{subfig:paris2}
      \begin{minipage}{0.33\textwidth}
         \includegraphics[width=\textwidth]{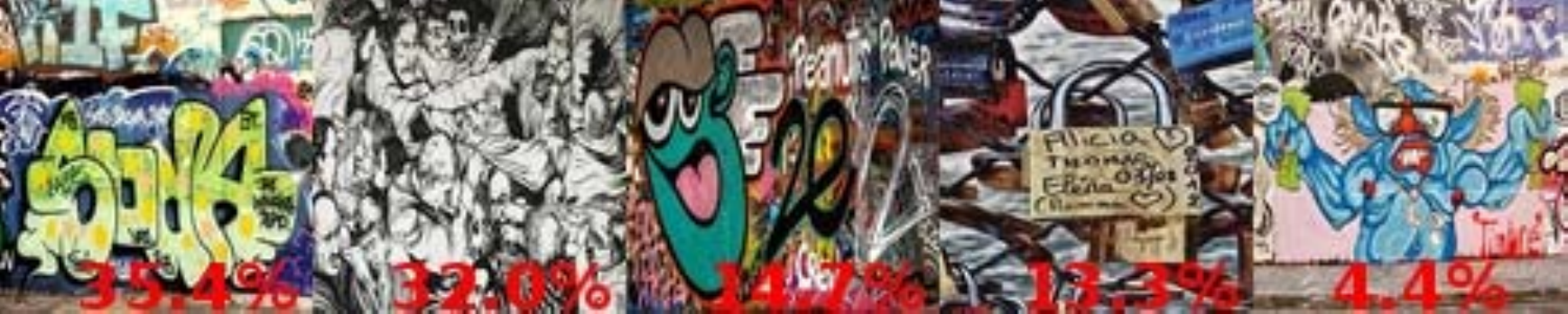} \newline
         \includegraphics[width=\textwidth]{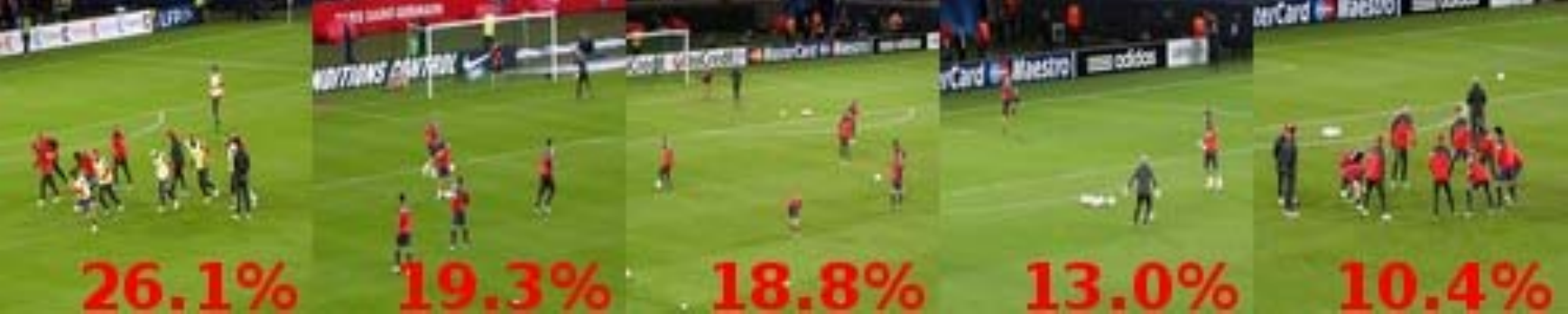} \\
         \includegraphics[width=\textwidth]{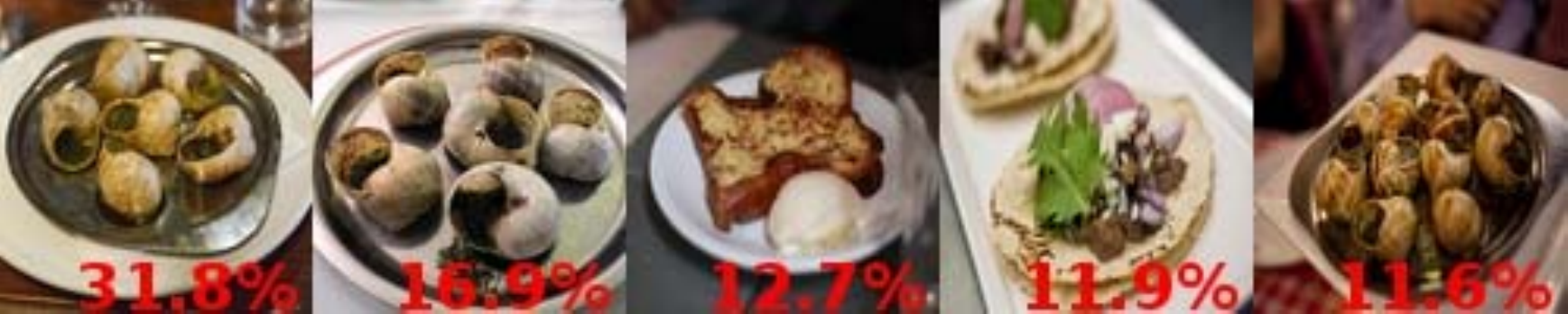} \\
         \includegraphics[width=\textwidth]{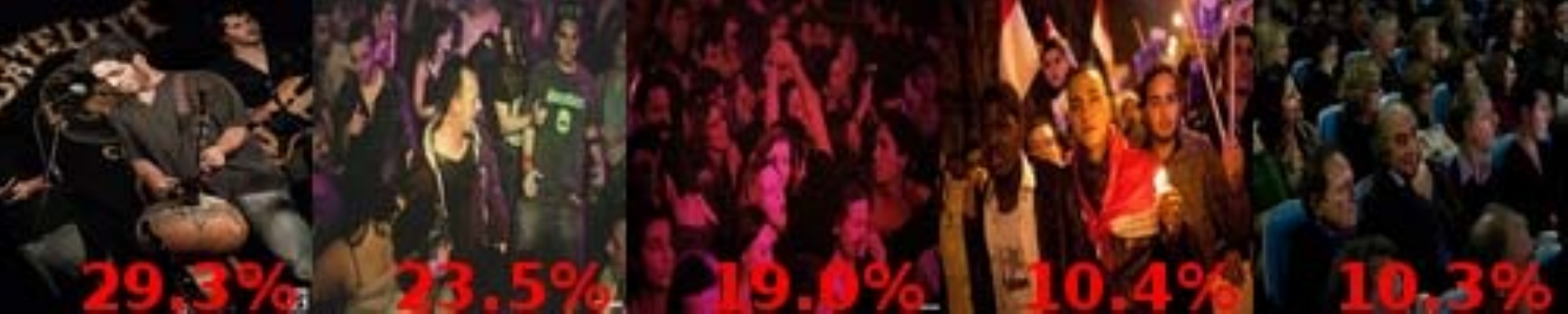} \\
         \includegraphics[width=\textwidth]{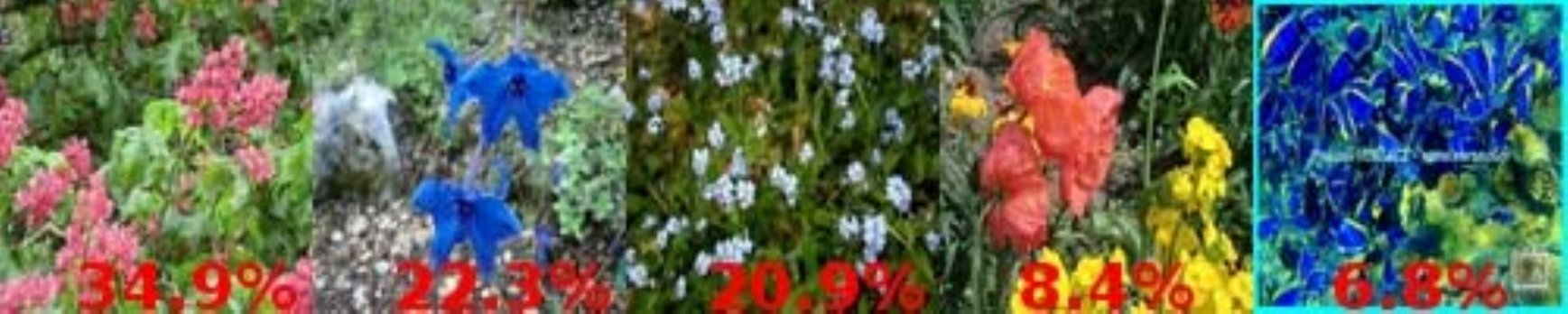} \\
         \includegraphics[width=\textwidth]{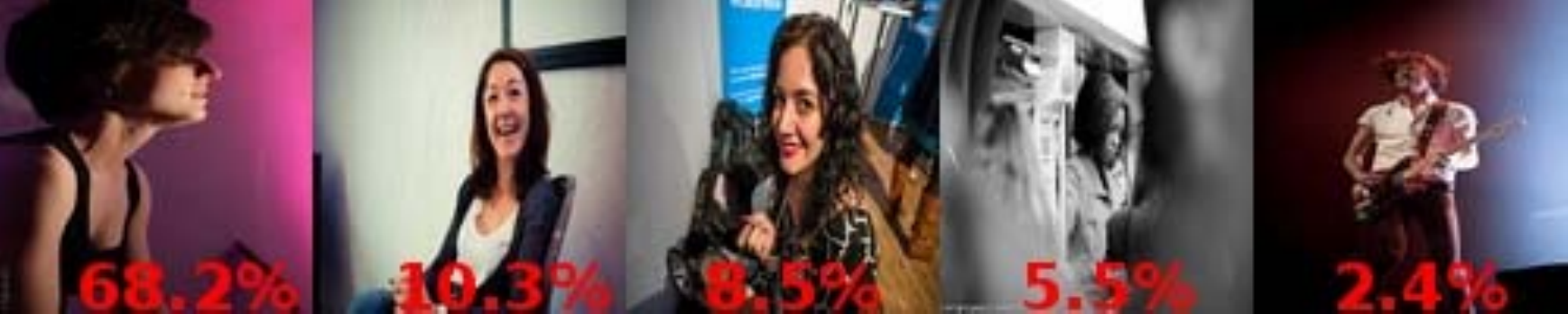} \\
         \includegraphics[width=\textwidth]{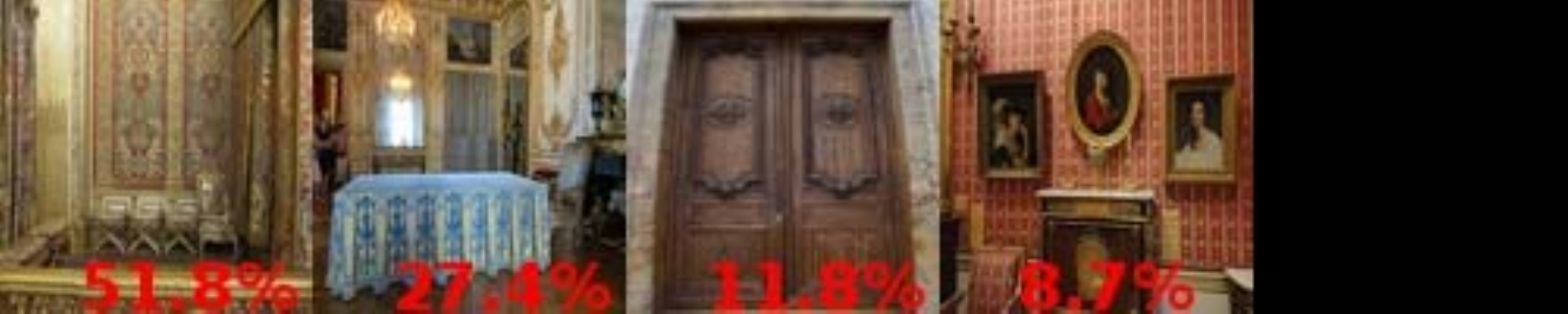} 
      \end{minipage}
   }
   \subfloat[Archetypes representing scene composition.]{\label{subfig:paris3}
      \begin{minipage}{0.33\textwidth}
         \includegraphics[width=\textwidth]{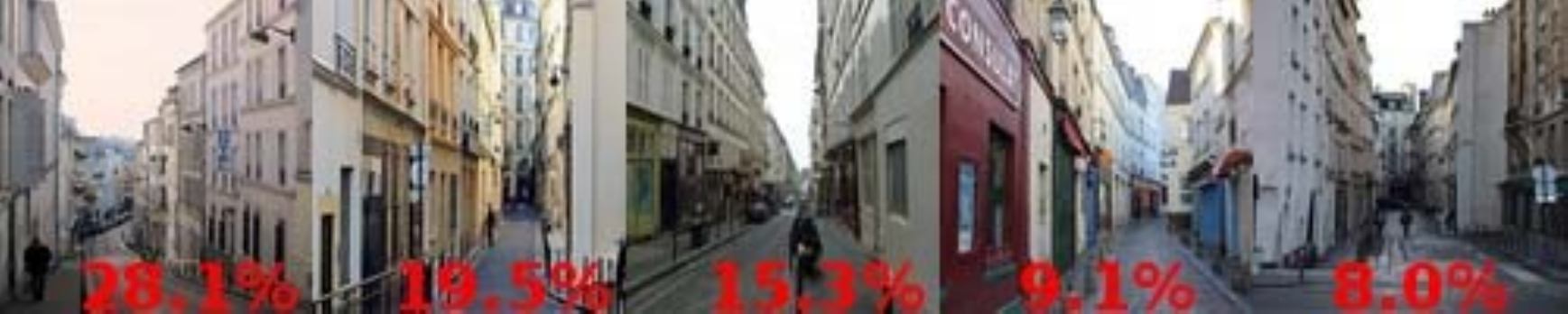} \newline
         \includegraphics[width=\textwidth]{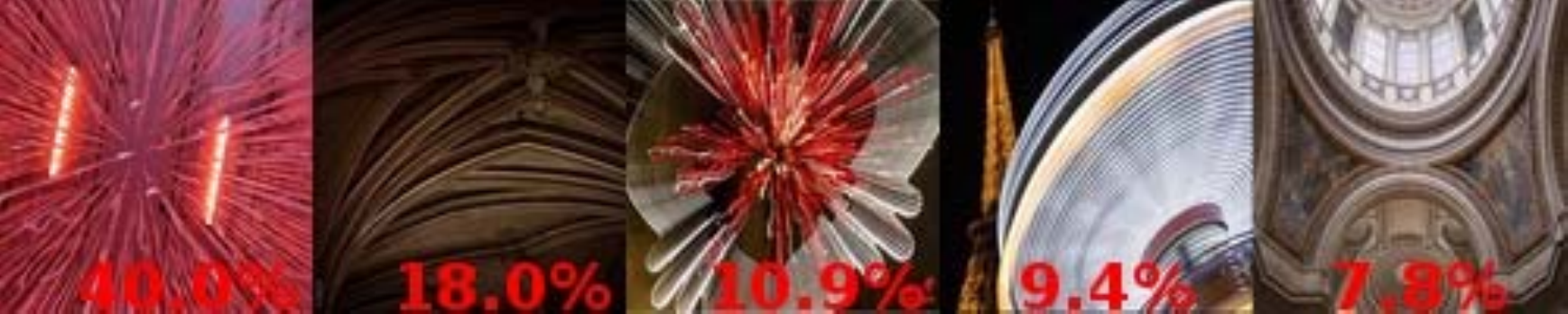} \\
         \includegraphics[width=\textwidth]{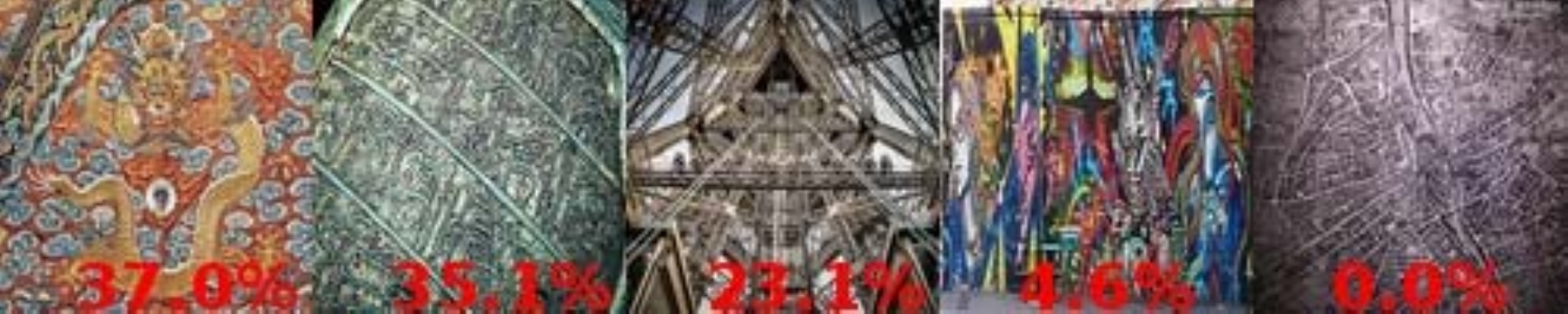} \\
         \includegraphics[width=\textwidth]{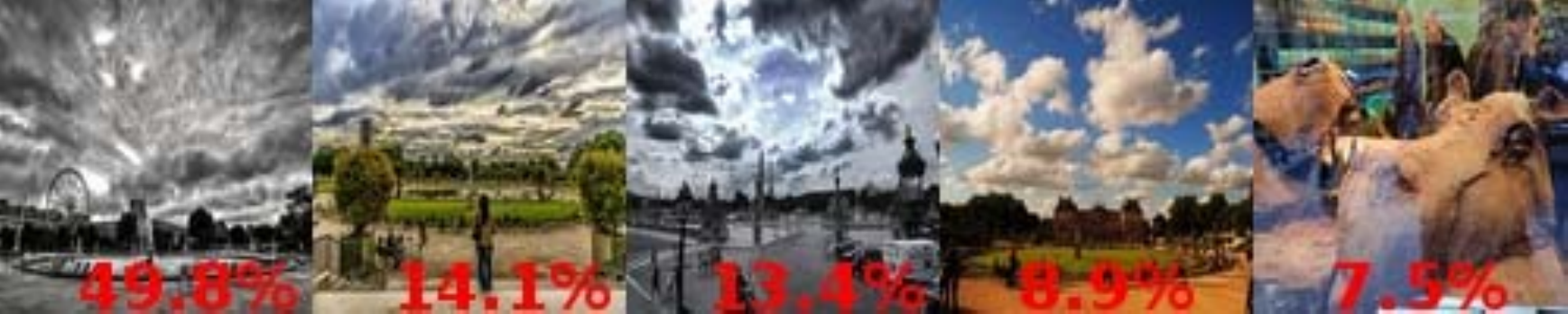} \\
         \includegraphics[width=\textwidth]{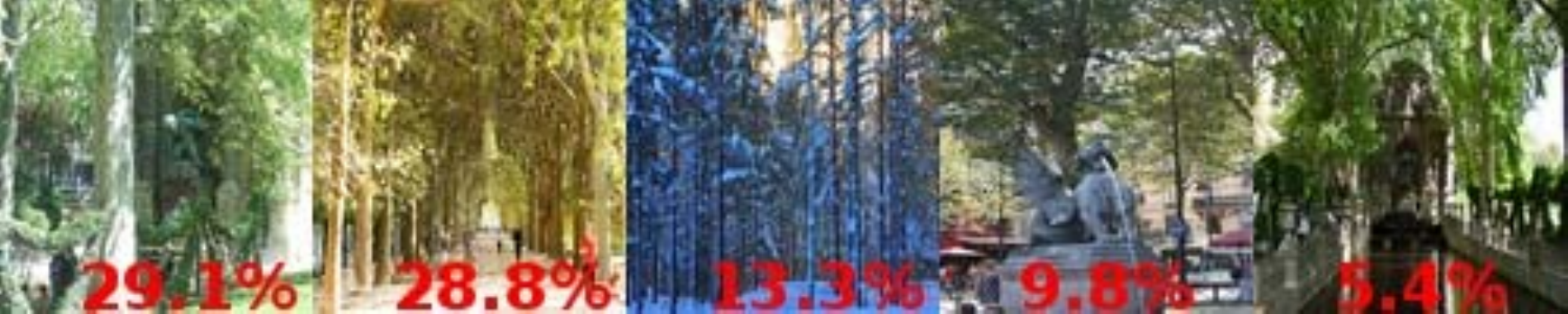} \\
         \includegraphics[width=\textwidth]{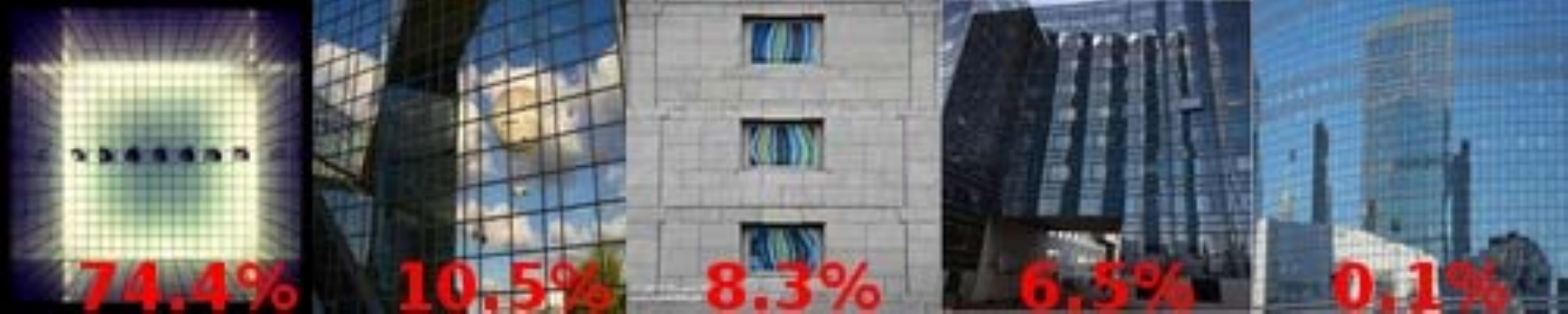}  \\
         \includegraphics[width=\textwidth]{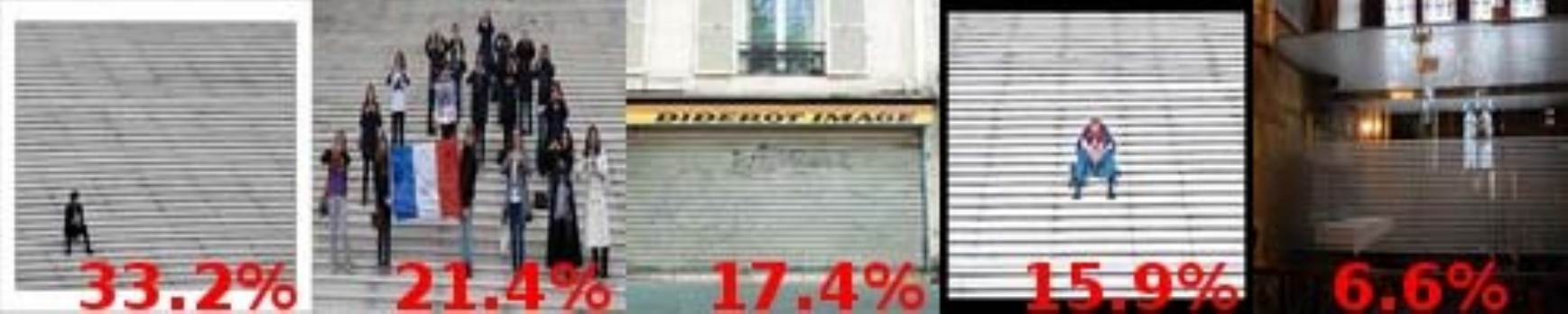} 
      \end{minipage}
   }
   \caption{Some archetypes learned from $36\,600$ pictures corresponding to the request ``Paris'' downloaded from Flickr.}
   \label{fig:paris_n}
\end{figure*}

Even though archetypes are learned in the space of Fisher vectors, which
are not displayable, each archetype can be interpreted as a sparse convex
combination of data points.  In Figure~\ref{fig:paris_n} we represent some of
the archetypes learned by our approach; each one is represented by a few
training images with some proportions indicated in red (the value of the
$\betab_j$'s). Classical landmarks appear in Figure~\ref{subfig:paris1}, which
is not surprising since Flickr contains a large number of vacation pictures. In
Figure~\ref{subfig:paris2}, we display several archetypes that we did not
expect, including ones about soccer, graffitis, food, flowers, and social gatherings.
In Figure~\ref{subfig:paris3}, some archetypes do not seem to have some
semantic meaning, but they capture some scene composition or texture that are
common in the dataset.
We present the rest of the archetypes in the supplementary material, and 
results obtained for other requests, such as London or Berlin.

\begin{figure*}[hbtp]
   \subfloat[Image with trivial interpretation.]{\label{subfig:paris1b}
      \includegraphics[width=0.32\textwidth]{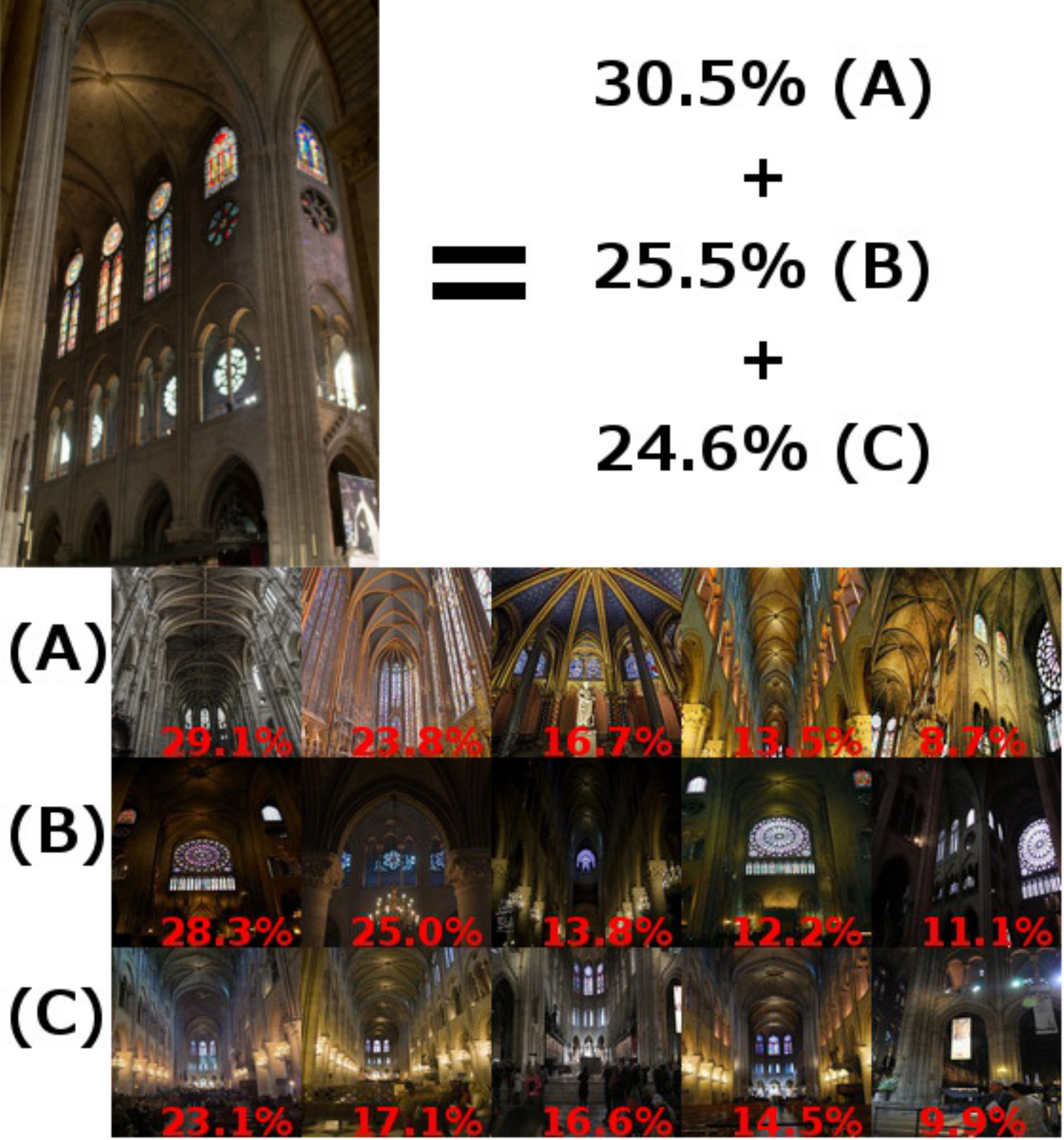} 
   } \hfill
   \subfloat[Image with interesting interpretation.]{\label{subfig:paris2b}
      \includegraphics[width=0.32\textwidth]{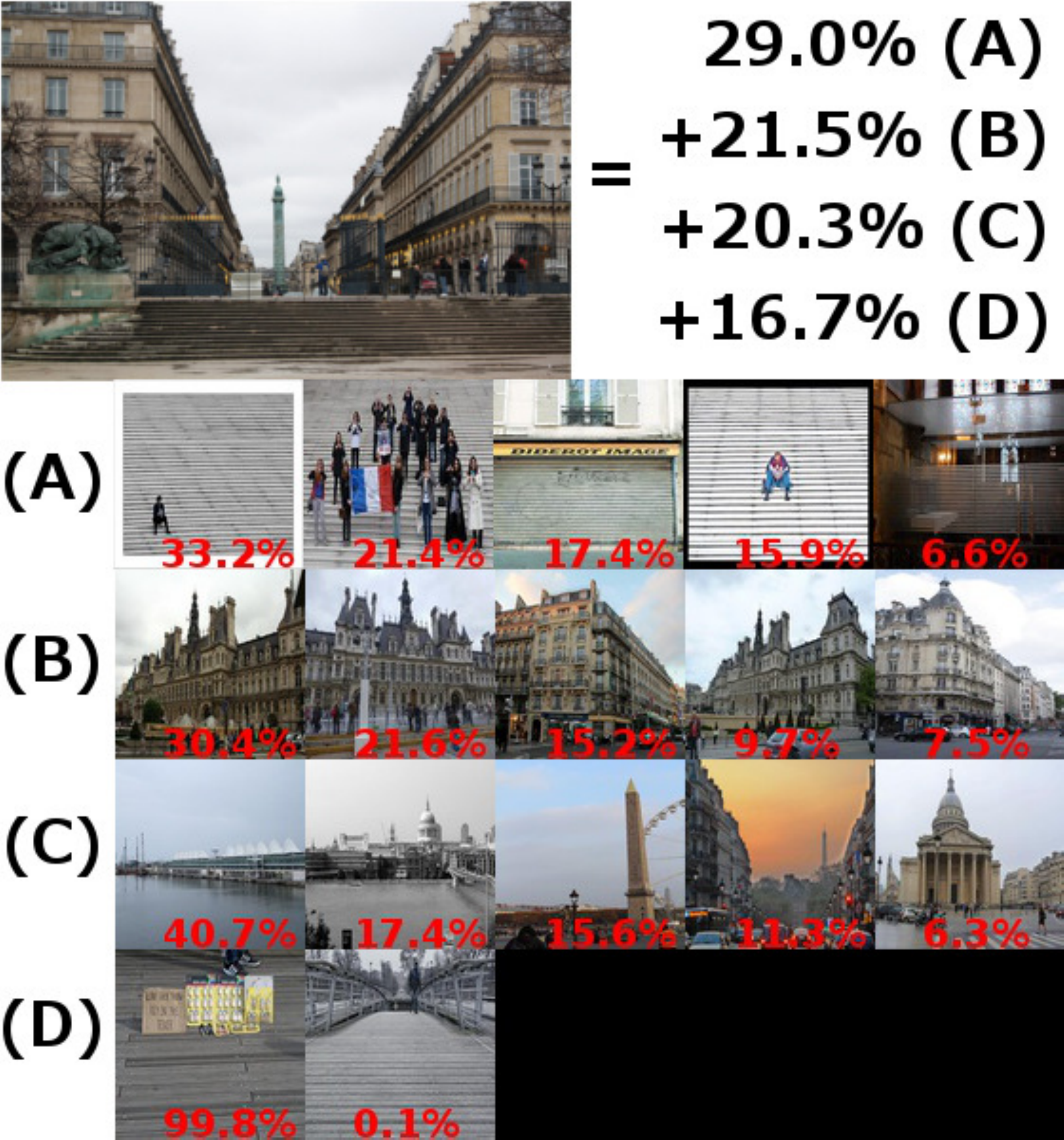} 
   } \hfill
   \subfloat[Image with bad (non-sparse) reconstruction.]{\label{subfig:paris3b}
      \includegraphics[width=0.32\textwidth]{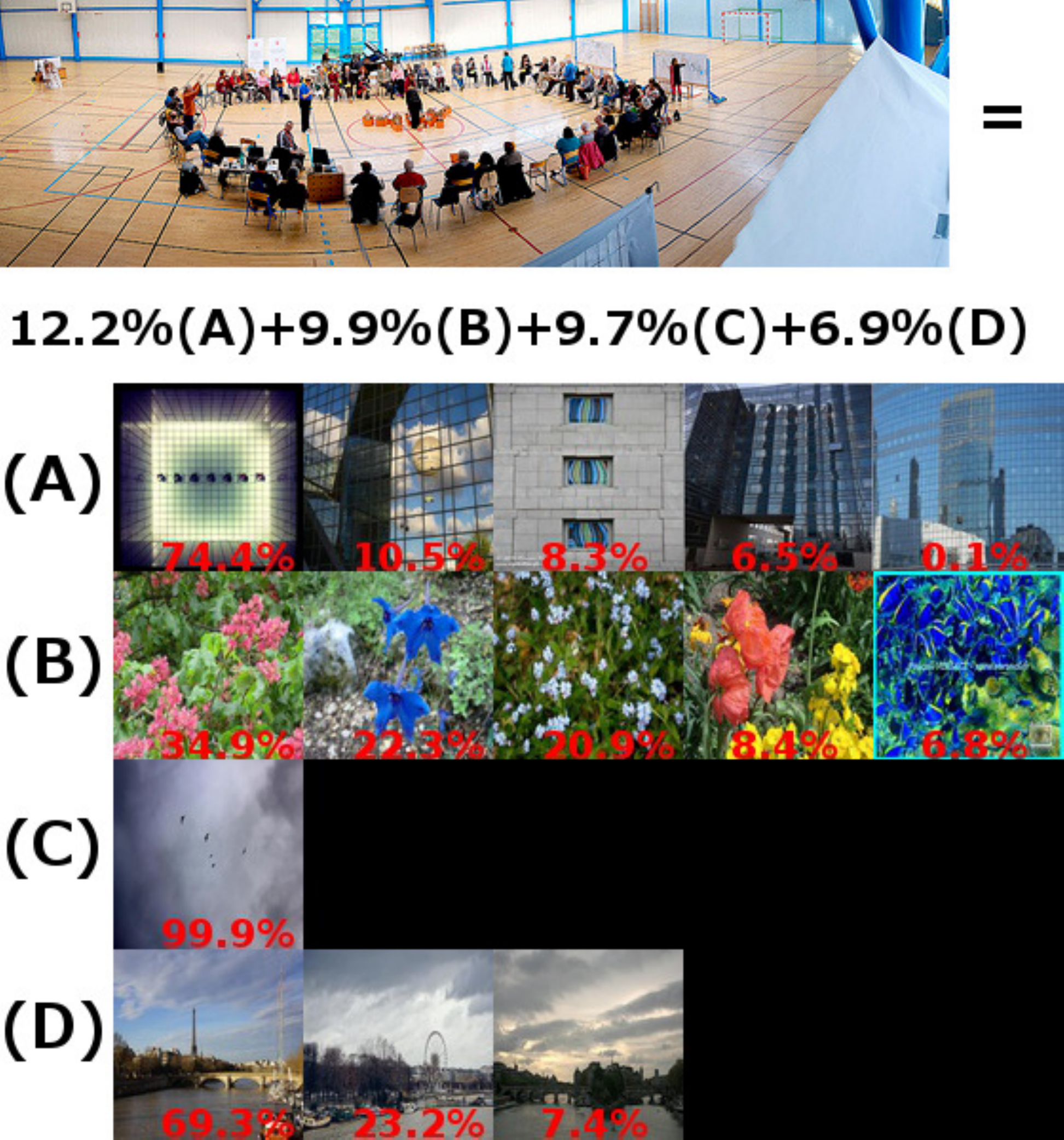} 
   } 
   \caption{Decomposition of some images onto the learned archetypal set.}
   \label{fig:paris_n2}
\end{figure*}

In Figure~\ref{fig:paris_n2}, we exploit the symmetrical relation between data
and archetypes. We show for four images how they decompose onto archetypes,
indicating the values of the~$\alphab_i$'s.  Some decompositions are trivial
(Figure~\ref{subfig:paris1b}); some others with high mean squared error are
badly represented by the archetypes (Figure~\ref{subfig:paris3b}); some others
exhibit interesting relations between some ``texture'' and ``architecture''
archetypes (Figure~\ref{subfig:paris2b}).

\section{Discussions}\label{sec:ccl}
In this paper, we present an efficient active-set strategy for
archetypal analysis.  By providing the first scalable open-source
implementation of this powerful unsupervised learning technique, we hope that
our work will be useful for applying archetypal analysis to various scientific
problems.  In particular, we have shown that it has promising applications in
computer vision, where it performs as well as sparse coding for prediction
tasks, and provides an intuitive visualization technique for large databases of
natural images. We also propose a robust version, which
is useful for processing datasets containing noise, or outliers, or both.

\subsection*{Acknowledgements}\label{sec:ack}
This work was supported by the INRIA-UC Berkeley Associated Team ``Hyperion'', a grant from the France-Berkeley fund, the project ``Gargantua'' funded by the program Mastodons-CNRS, and the Microsoft Research-Inria joint~centre.  
{\small
\bibliographystyle{ieee}
\bibliography{ref}
}

\end{document}